\documentclass[11pt]{article}
\usepackage[a4paper,margin=2.5cm]{geometry}
\usepackage[T1]{fontenc}
\usepackage{lmodern}
\usepackage{microtype}
\usepackage{amsmath,amssymb}
\usepackage{graphicx}
\usepackage{pdfpages}
\usepackage{booktabs}
\usepackage{array}
\usepackage{xcolor}
\usepackage{caption}
\usepackage{enumitem}
\usepackage{setspace}
\usepackage[numbers,super,sort&compress]{natbib}
\usepackage[colorlinks=true,linkcolor=blue!50!black,citecolor=blue!50!black,urlcolor=blue!50!black]{hyperref}
\setlist{nosep,leftmargin=1.4em}
\newcommand{\fmt}[1]{{\normalfont\scshape #1}}
\newcommand{\cho}{\fmt{choose}}
\newcommand{\ans}{\fmt{answer}}
\newcommand{\jsn}{\fmt{json}}
\newcommand{\twc}{\fmt{twice}}

\title{Language model capabilities fail in a shared, predictable order under compression and perturbation}
\author{Zhixiang Wang and Ulas Bagci\\[2pt]\small Department of Radiology, Feinberg School of Medicine, Northwestern University, Chicago, IL, USA\\\small Correspondence: \texttt{zhixiang.wang@northwestern.edu}}
\date{}

\begin{document}
\maketitle

\begin{abstract}
\noindent Compression can preserve a language model's average accuracy while selectively degrading individual capabilities. Whether these losses follow method-specific patterns or a transferable order is unclear. We applied seven perturbation and compression methods to eight instruction-tuned models, scoring 16 tasks in four answer formats against preregistered predictions. Short-answer capabilities failed in nearly the same order under four random perturbations with different mechanisms (median correlation 0.90) and, within each model, under pruning and quantization (0.875). Intact answer margin, answer frequency and answer format predicted this order. Fragility profiles measured under random perturbation predicted which tasks and facts compression removes first and, after four-task calibration, held-out task accuracy (median absolute error 0.11--0.12, against 0.20--0.21 for uniform decline); profiles borrowed from other model families did almost as well. The order did not transfer across models under GPTQ quantization or hold for long-form tasks. Within short answers, what compression removes is predictable before deployment.
\end{abstract}

Language models are routinely pruned and quantized before deployment\citep{zhu2024survey,frantar2023sparsegpt,sun2024wanda,frantar2023gptq}, and are further damaged by faulty hardware\citep{huang2026errors,sun2025demystifying} and by layer removal\citep{men2024shortgptarxiv,gromov2024unreasonable}. Such damage rarely shows in an aggregate score: a compressed model can lose a few points on average while some of its capabilities collapse\citep{hooker2019compressed}. Whoever deploys a damaged model needs to know what it loses first.

Studies that have measured this disagree. Pruning has been reported to impair closed-book fact recall before in-context abilities\citep{jaiswal2024compressing}, a unified evaluation of pruning, quantization and distillation finds knowledge preserved while reasoning and instruction following degrade\citep{vonrad2026unicomp}, and a study of eleven compression methods found knowledge about obscure entities spared relative to popular ones\citep{wu2026asymmetric}. These studies vary the damage, the tasks and the scoring rule at once: the task reported as fragile is usually generated and scored by exact match, the task reported as robust usually multiple choice with a chance floor of 25--50\%, and format and scoring alone can change how capable a model appears\citep{schaeffer2023mirage,tam2024speak,wang2024my}. Without an account of what sets the order, every new method of compression is profiled from scratch.

The usual expectation is that the damage decides: compression methods choose what to remove by its effect on calibration text\citep{frantar2023sparsegpt,sun2024wanda}, losses are attributed to particular methods\citep{li2024evaluating,marchisio2024multilingual}, and lesions confined to particular components of language models produce deficits that depend on what was lesioned\citep{roll2026aphasias}. Neuropsychology suggests otherwise. In semantic dementia, less familiar knowledge is lost before familiar knowledge and naming before recognition, and errors drift towards familiar answers\citep{rogers2004structure,patterson2007where}; connectionist networks reproduce these regularities wherever they are lesioned\citep{hinton1991lesioning,rogers2004structure}, suggesting that the order follows from what a network has learned rather than from where it is damaged. Whether this holds for language models, and whether the order can then be anticipated, is not known.

\begin{figure}[tbp]
  \centering
  \includegraphics[width=\linewidth,height=0.72\textheight,keepaspectratio]{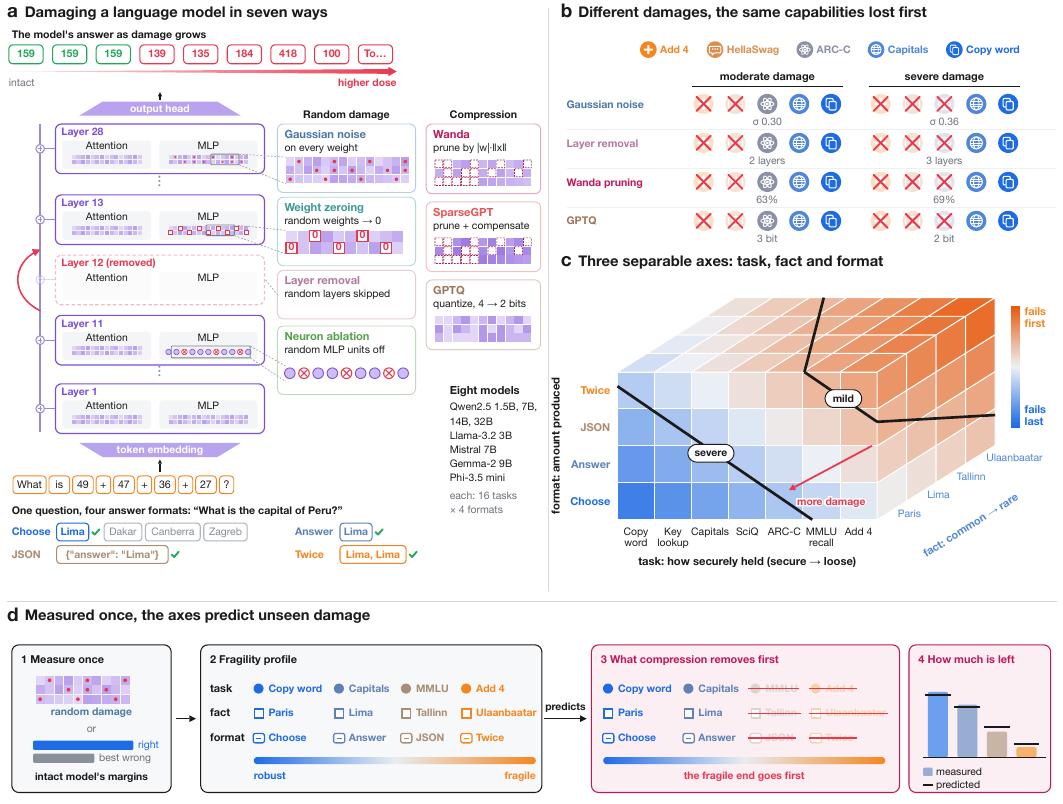}
  \caption{\textbf{Study design and main findings.} \textbf{a}, Seven damages at rising dose: four random damages (Gaussian noise, weight zeroing, neuron ablation, layer removal) and three compression methods (Wanda, SparseGPT, GPTQ), in eight models; each of 16 tasks is asked in four formats (\cho{}, \ans{}, \jsn{}, \twc{}). Top, the answer of Qwen2.5-7B to one item as the noise dose rises. \textbf{b}, Every damage removes the same capabilities first, each at its own dose. \textbf{c}, Three separable axes: how securely the task is held, how familiar the answer is and how much of it must be produced. Because the effects add, the failure front is a plane, and more damage moves it towards the robust corner; the fact axis was measured within recall tasks only. \textbf{d}, Measured once under random damage, in the model itself or in other models, the profile predicts what compression removes first and how much accuracy it leaves. Panels b--d are schematics.}
  \label{fig:overview}
\end{figure}

Here we separate the damage, the task, the familiarity of the fact and the answer format, applying seven damages, including the pruning and quantization methods used in practice, to eight instruction-tuned models and scoring 16 tasks in four answer formats against preregistered predictions (Fig.~\ref{fig:overview}). The damage sets how fast a model degrades; what fails first is set by three axes that do not change each other's order: how securely the intact model holds a task, how familiar the answer to a fact is and how much of the answer must be produced. Two follow the regularities of semantic dementia. A profile measured once predicts what unseen damage will take, down to single facts, and, after a brief calibration, the accuracy that remains, even without measuring the target model's own fragilities. What a damaged language model loses first is thus decided before the damage, by what it knows and how it must answer, and can be measured once rather than rediscovered for every new model and method of compression.

\section*{Results}

\subsection*{Damage sets the rate of decline, not the order}

For a measure $m$, one task asked in one answer format, with chance level $\kappa_m$ and intact accuracy $a_m(0)$, we define the chance-corrected score relative to the intact model, $c_m(d) = (a_m(d)-\kappa_m)/(a_m(0)-\kappa_m)$, at dose $d$. Within each cell, one model under one damage, we fit the rank-one model
\begin{equation}
  c_m(d) \approx \exp\!\big(-n_m\, h(d)\big),
  \label{eq:rank1}
\end{equation}
in which the damage contributes only a strength $h(d)$ shared by all measures, and each measure contributes its fragility exponent $n_m$ (Methods): the damage turns a single dial, and each measure fails at its own rate as the dial turns. The model fits well ($R^2$ 0.87--0.95 for noise, zeroing and neuron ablation, 0.74--0.88 for random layer removal and 0.70--0.93 for the compression methods; Extended Data Table~1). Two damages therefore share an order exactly when their exponents correlate, whatever their doses mean. We split each cell's log exponents into a task effect, how early a task fails relative to the others, and a format effect (equation~(\ref{eq:additive}), Methods), and correlated task effects between cells.

\begin{figure}[tbp]
  \centering
  \includegraphics[width=\linewidth,height=0.72\textheight,keepaspectratio]{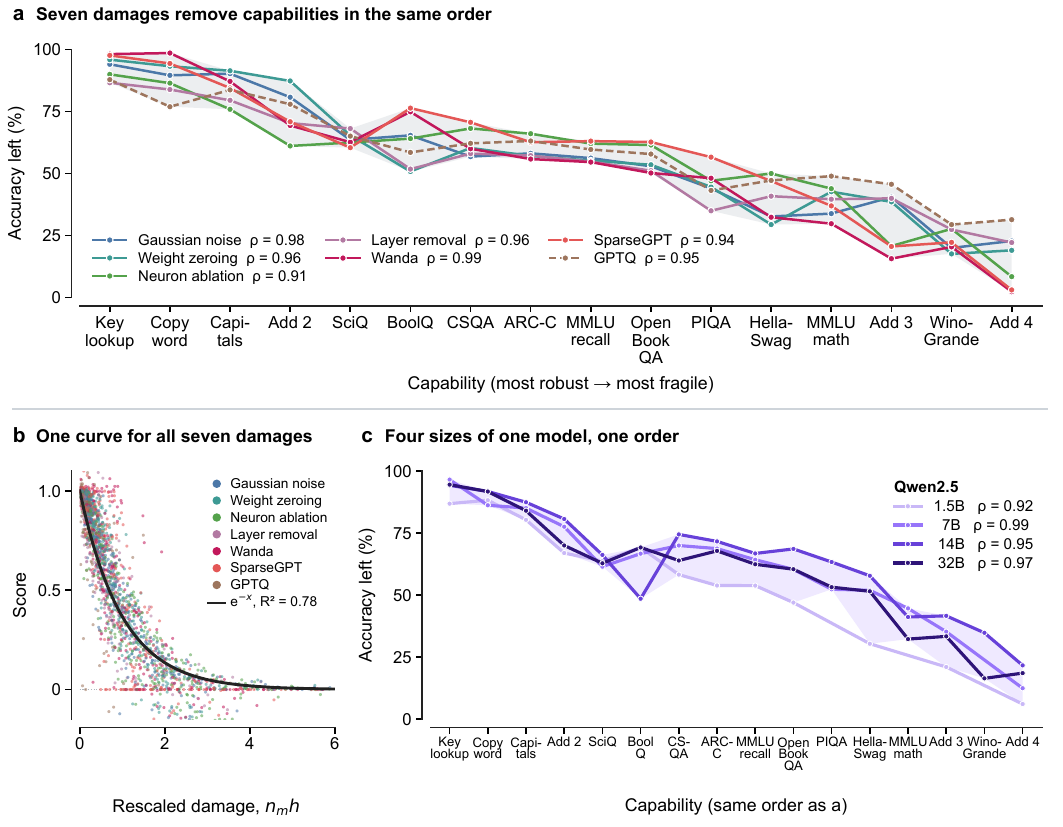}
  \caption{\textbf{Damage sets the rate, not the order.} \textbf{a}, Accuracy left in 16 tasks (\ans{}, chance-corrected) at the dose that halves each model's overall ability, mean of eight models, for each damage; tasks from most robust to most fragile; grey band, range over damages; $\rho$, Spearman correlation with the mean order. \textbf{b}, Relative score of every measure of Qwen2.5-7B against $n_m h(d)$ from one joint fit for all damages, with $e^{-x}$ ($R^2=0.78$; eight models 0.72--0.82; exploratory). \textbf{c}, As in \textbf{a}, averaged over damages, for four sizes of Qwen2.5; band, range over sizes; $\rho$ against the mean over sizes.}
  \label{fig:order}
\end{figure}

Four random damages that act through different mechanisms make the tasks fail in the same order (Fig.~\ref{fig:overview}b). Within a model, the task effects of two random damages correlate with a median of 0.904 over 35 damage pairs in seven models (95\% interval 0.81--0.95; Fig.~\ref{fig:order}a), and across models, under the same damage, with medians of 0.832--0.913 (69 pairs); both preregistered predictions hold, and correcting for the disagreement between two seeds of the same damage raises the agreement between damages to 0.92--0.94 (Extended Data Fig.~1). Importance-guided compression, which decides what to remove by its effect on calibration text, could have rewritten the order. It did not: task effects under Wanda, SparseGPT and GPTQ correlate with the same model's random-damage task effects with a median of 0.875 over 18 cells (preregistered threshold 0.7), and across models with medians of 0.924 (Wanda) and 0.927 (SparseGPT). Without the three most robust and the most fragile task, the agreement remains 0.727, against a 95th percentile of 0.184 with shuffled labels. The within-model agreement remains 0.86--0.88 with stricter fitting windows, and the across-model agreement 0.87--0.93 with one model per family (exploratory; Supplementary Note~9). GPTQ follows the order within a model (0.53--0.90) but not across models (median 0.095), so the preregistered across-model prediction fails because of GPTQ alone (Extended Data Fig.~1).

A joint fit with one exponent per measure places all seven damages of a model on a single curve (72--82\% of the variance against 84--91\% for separate fits; exploratory; Fig.~\ref{fig:order}b). The order holds at 32 billion parameters (the random damages agree at 0.843 in Qwen2.5-32B, and its order agrees with the smaller Qwen models at 0.898; Fig.~\ref{fig:order}c) and, less precisely, item by item (preregistered, partial; Extended Data Fig.~1). It does not extend to long answers: on four long-form tasks (reasoning\citep{cobbe2021gsm8k}, code\citep{chen2021humaneval,austin2021mbpp}, instruction following\citep{zhou2023ifeval} and function calling\citep{patil2025bfcl}) in four models, 10 of 16 task--model pairs kept their position, short of the preregistered 75\%, and multi-step reasoning became more fragile under pruning (Extended Data Fig.~2). Our claims concern short answers.

\subsection*{The order has three separable axes: task, fact and format}

\begin{figure}[tbp]
  \centering
  \includegraphics[width=\linewidth,height=0.72\textheight,keepaspectratio]{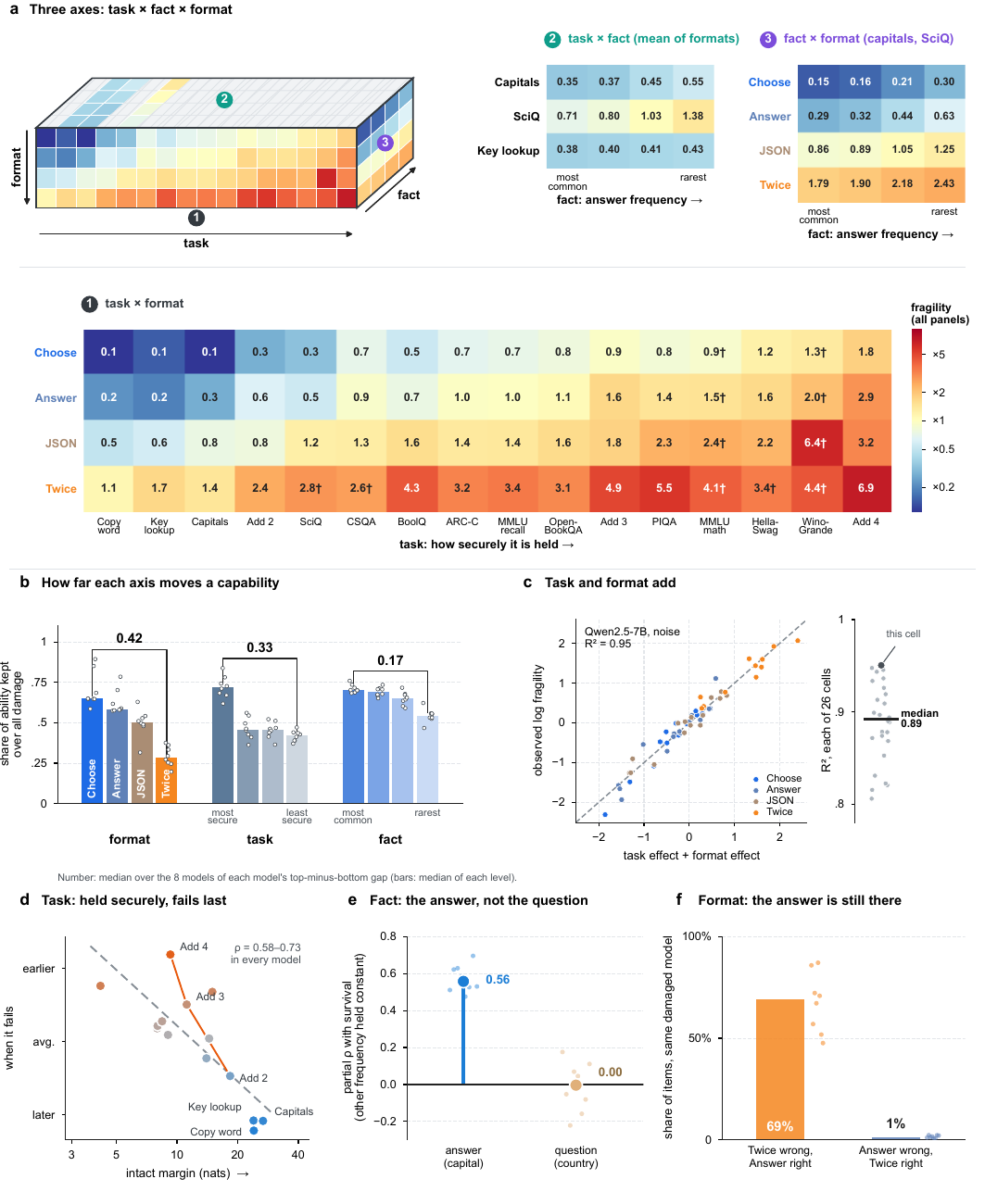}
  \caption{\textbf{The three axes add.} \textbf{a}, Fragility (fold change of the exponent, averaged over models and random damages) on three faces of the task $\times$ fact $\times$ format space: (1) task $\times$ format, 16 tasks; (2) task $\times$ fact, quarters of answer frequency; (3) fact $\times$ format, capitals and SciQ. $\dagger$, measurable in fewer than four models. \textbf{b}, How far each axis moves a capability: share of ability kept over all damage, first against last quarter of each axis; numbers, median gap over eight models; dots, models (exploratory). \textbf{c}, Task plus format effects against observed log exponents (Qwen2.5-7B, noise, $R^2=0.95$); right, $R^2$ in all 26 random-damage cells (median 0.89). \textbf{d}, Tasks the intact model holds securely fail later (Spearman 0.58--0.73 in every model). \textbf{e}, Partial correlation of capital survival with answer frequency (country held constant) and with country frequency (capital held constant). \textbf{f}, Items a damaged model fails in \twc{} but answers in \ans{}, and the reverse.}
  \label{fig:axes}
\end{figure}

We call an axis a property of what is asked that sets when a capability fails, whatever the damage. The task and the format act on whole capabilities; the fact acts within a task, deciding which facts are lost first (Fig.~\ref{fig:axes}a).

On one scale, the share of ability kept over all damaged conditions, the first and last quarters of each axis differ by 0.42 for the format, 0.33 for the task and 0.17 for the familiarity of facts (medians over models; exploratory; Fig.~\ref{fig:axes}b).

The three axes combine simply. If fragility were a property of each capability alone, changing the answer format would reshuffle the tasks; it does not. Within a model and damage, a task effect plus a format effect explains a median of 89\% of the variance of the log exponents (81--95\% over 26 cells; Fig.~\ref{fig:axes}c), and the formats keep their order for every task in 25 of 26 random-damage cells and 19 of 21 compression cells. Rare answers fail first whichever way the answer must be written: the frequency effect (defined below) had medians of 0.40 in \ans{}, 0.37 in \jsn{} and 0.21 in \twc{}, positive in every model--task cell. Conversely, among items the intact model answered correctly in all three formats, \ans{} outlasted \jsn{} and \jsn{} outlasted \twc{} among rare and among common answers alike, in all 11 testable cells (both preregistered). The separation is approximate: the frequency effect halves in \twc{}, where template failures mask what the model knows, and the fact axis was measured in two recall tasks.

Along the task axis, what the intact model holds less securely fails first. On intact models only, we measured the answer margin of each task, the mean log-probability gap between the correct candidate and the best distractor. Tasks with a smaller margin fail earlier in every model (Spearman correlation 0.58--0.73; Fig.~\ref{fig:axes}d), and the margin explains a median of 63\% of the variance of the damage-averaged task effects. This is not merely that harder tasks fail first: intact accuracy is the weaker predictor in six of seven models, and regressing out the margin lowers the median between-model correlation of task effects from 0.946 to 0.678, whereas regressing out intact accuracy leaves 0.921. The margin matters for a simple reason. If an item survives while its margin exceeds the perturbation that damage adds to it, survival depends only on the ratio of the two, and with the rank-one form this implies a power law, $n_t\propto\mu_t^{-\beta}$, for a task with typical margin $\mu_t$ (Supplementary Note~10). Across the eight models, task effects fell with the log margin (median Pearson $-0.76$, negative in every model) with $\beta$ near 2 (median 1.95), the value expected if damage strength grows with the variance of the perturbation (exploratory). The number of steps also places a task on this axis: every added addend makes arithmetic more fragile (20 of 20 cells). Knowledge is not one point on this axis: capitals of well-known countries have one of the four largest margins in every model and survive longest, whereas MMLU knowledge questions, with median margins, are more fragile than the average task (Extended Data Fig.~3).

Within a task, facts whose answers are rare fail first. Intact models already answer rare facts less accurately\citep{kandpal2023large}; we asked whether damage removes them first. We counted each answer in the open pretraining corpus Dolma\citep{soldaini2024dolma} with an exact $n$-gram index\citep{liu2024infinigram} and correlated the count with each item's survival, its mean accuracy over all damaged conditions, among items the intact model answered correctly (Methods). For capitals, the co-occurrence of country and capital predicted survival in all eight models (median Spearman correlation 0.42, range 0.34--0.55); for SciQ, whose answers are written without options, the frequency of the answer did the same (0.38, 0.30--0.45; both preregistered; Extended Data Fig.~4), also with answer length or item margin held constant (partial correlations 0.26--0.55). For the answers of the key-lookup task, codes given in the prompt that need not be recalled, frequency barely predicted survival (0.08; preregistered control). The effect held under each of the seven damages (0.23--0.41): pruning and quantization remove the long tail of what a model knows before its common core. Among capitals, the frequency of the capital predicted survival with the frequency of the country held constant (partial correlation 0.48--0.70), but not the reverse ($-0.22$ to $0.18$; Fig.~\ref{fig:axes}e), which explains why grouping facts by the popularity of the entity asked about reached the opposite conclusion (Discussion).

The errors themselves move towards familiar answers, as in patients (exploratory; Extended Data Fig.~3): a wrong capital was the more frequent of the two in 62--100\% of cases, the most frequent distractor drew 48--62\% of choice errors against 33\% by chance, and yes/no answers drifted towards the more common answer.

Along the format axis, producing fails before recognizing. Choosing among candidates asks the model only to rank a few answers, as word--picture matching asks a patient; writing the bare answer requires producing it exactly, as naming does; templates add structure that must also be exact. Relative to \cho{}, the median exponent is 1.5, 2.5 and 8 times larger for \ans{}, \jsn{} and \twc{} under random damage, and 1.4, 2.1 and 6.5 times under compression (Extended Data Fig.~3). Scoring only whether the answer is present, whatever the template, separates the two parts of this axis: the exponent is still 1.45 times larger for \ans{} than for \cho{} in every model, and 1.64 and 1.93 times for \jsn{} and \twc{} (medians, random damage; preregistered analysis). Producing an answer is thus more fragile than choosing it in content\citep{wang2024my}, and most of the extra fragility of the templated formats is the template: at intermediate damage, 83\% of \twc{} errors violate the template, against 0.2\% of \ans{} errors, 79\% of these still contain the correct answer, and on items that \twc{} gets wrong the same model answers correctly in \ans{} 69\% of the time, against 1--4\% for the reverse (Fig.~\ref{fig:axes}f). Function calling behaves the same way\citep{tam2024speak,patil2025bfcl}: 68.4\% of damage-induced errors were format failures, 77.4\% of these with the right function and arguments, and strict scoring made the task 2.3 times more fragile than content scoring (preregistered; Extended Data Fig.~3).

Writing an answer twice is the exception: \jsn{}, whose template has more tokens that must be exact, should fail before \twc{}, but \twc{} fails first. Four preregistered explanations based on the intact model's confidence in what it must write did not separate the two (Supplementary Note~2; Supplementary Table~3). What fails is producing the answer inside the template, and why remains open.

\subsection*{Measured once, the axes predict unseen damage}

\begin{figure}[tbp]
  \centering
  \includegraphics[width=\linewidth,height=0.72\textheight,keepaspectratio]{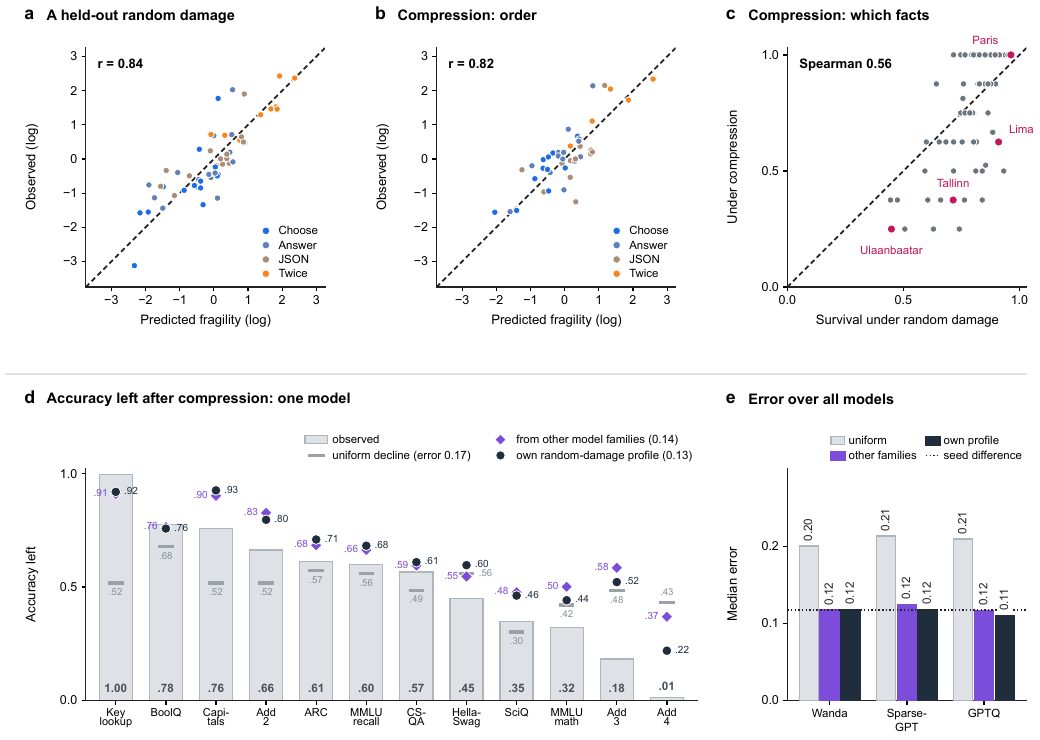}
  \caption{\textbf{Measured once, the axes predict unseen damage.} \textbf{a}--\textbf{d}, Qwen2.5-7B. \textbf{a}, Log exponents of neuron ablation predicted from the other three random damages. \textbf{b}, Wanda exponents predicted from all four random damages. \textbf{c}, Survival of each capital under Wanda against its survival under random damage. \textbf{d}, Accuracy (fraction correct, not corrected for chance) after Wanda pruning to 63\% sparsity, in \ans{}, for the twelve tasks held out from calibration. Bars and the numbers at their base, measured accuracy; markers and the numbers beside them, predictions from the model's own random-damage profile (circles), from the profile transferred from other model families, without the model's own profile (diamonds) and from a uniform decline (dashes). The strength of the damage was estimated on the four calibration tasks (WinoGrande, OpenBookQA, PIQA, copying a word); numbers in the key, mean absolute error over the twelve tasks. \textbf{e}, Median error over all models for each compression method, for the uniform decline (grey), the transferred profile (purple) and the model's own profile (dark); dotted line, the difference between two seeds of the same random damage (exploratory). All predictions computed in preregistered analyses; summary statistics over models in the main text.}
  \label{fig:predict}
\end{figure}

We tested at five levels of increasing demand whether a profile, the set of fragility exponents of a model's capabilities, predicts unseen damage (Table~\ref{tab:predict}). The first three predict an order and use no data from the damage being predicted; the last two predict an amount and need one number from the compressed model, the strength of the damage, estimated on four fixed calibration tasks and never on the task being predicted. The order is thus predicted entirely by transfer, whereas the amount requires a limited measurement of the compressed model. Fitted on three random damages, task and format effects predict the log exponents under the fourth with a median correlation of 0.851, against 0.597 and 0.512 for each effect alone (20 of 20 cases); fitted on all four, they predict the exponents under compression at 0.807, beating both single effects in 18 of 19 cells (Fig.~\ref{fig:predict}a,b). The survival of a fact under random damage predicted its survival under compression (median Spearman correlations 0.69 for Wanda, 0.71 for SparseGPT and 0.58 for GPTQ, positive in all 48 cells; preregistered; Fig.~\ref{fig:predict}c), better than answer frequency (0.23--0.37) or the intact item margin (0.09--0.13), and with partial correlations of 0.51--0.63 when both are held constant (preregistered).

\begin{table}[tbp]
\centering
\small
\caption{\textbf{Five levels of prediction, from the order of capabilities to the accuracy they keep.} Levels 1--3 predict an order and use no data from the predicted damage. Levels 4 and 5 predict accuracy and take one strength per compression condition from four fixed calibration tasks of the compressed model; every measure of the other twelve tasks is predicted with all its formats held out. Values are medians over cells (model and damage, or model, task and compression method). As a reference for levels 4 and 5, the accuracy of the same measure under two runs of the same random damage at the same dose differs by a median of 0.117 (exploratory).}
\label{tab:predict}
\renewcommand{\arraystretch}{1.25}
\begin{tabular}{@{}>{\raggedright\arraybackslash}p{0.5cm}>{\raggedright\arraybackslash}p{3.0cm}>{\raggedright\arraybackslash}p{3.4cm}>{\raggedright\arraybackslash}p{2.6cm}>{\raggedright\arraybackslash}p{3.2cm}>{\raggedright\arraybackslash}p{1.7cm}@{}}
\toprule
Level & Predicted & From & Result & Against & Cells \\
\midrule
1 & Order of capabilities under a held-out random damage & task and format effects of the other three random damages & $r=0.851$ & task effects alone 0.597; format effects alone 0.512 & better in 20 of 20 \\
2 & Order of capabilities under Wanda, SparseGPT and GPTQ & task and format effects of the four random damages & $r=0.807$ & task effects alone 0.525; format effects alone 0.575 & better in 18 of 19 \\
3 & Which facts compression removes (capitals, SciQ) & survival of each fact under the four random damages & Spearman 0.69, 0.71, 0.58 (Wanda, SparseGPT, GPTQ) & answer frequency 0.36, 0.37, 0.23; intact item margin 0.09--0.13; key-lookup control 0.12--0.13 & positive in 48 of 48 \\
4 & Accuracy left in each capability after compression & the model's own random-damage exponents; strength from four calibration tasks & absolute error 0.110--0.118 & uniform decline 0.200--0.213 & better in 23 of 23; bound of 0.10 not met \\
5 & The same, without the target model's own fragilities & exponents of the same task and format in other model families; same calibration & absolute error 0.117--0.125 & uniform 0.200--0.213; intact margins 0.133--0.142; intact accuracy 0.185--0.205; format alone 0.171--0.175 & better than uniform in 23 of 23 \\
\bottomrule
\end{tabular}
\par\smallskip\raggedright\footnotesize All five analyses were preregistered (Supplementary Table~2: T3, T10, X1--X2, A1--A2, B1, R1--R4).
\end{table}

Correlations predict an order, not an amount. Because the damage enters only through one strength, that strength, estimated on the compressed model, together with exponents measured under random damage, should give the accuracy left in each capability. We fixed four calibration tasks in advance, chosen from the intact models' margins to span fragile and robust tasks (WinoGrande, OpenBookQA, PIQA and copying a word), estimated the strength of each informative compression condition from them, and predicted every measure of the other twelve tasks, each held out with all its formats (Methods). The median absolute error was 0.118 for Wanda, 0.118 for SparseGPT and 0.110 for GPTQ, against 0.200--0.213 for a uniform decline, in which every capability loses the same fraction; the profile was more accurate in all 23 model--method cells, with 44--46\% smaller error (preregistered; Fig.~\ref{fig:predict}d,e; Extended Data Fig.~5). The error missed our preregistered bound of 0.10 but sits at the noise ceiling, the difference between two seeds of the same random damage (0.117; exploratory; Fig.~\ref{fig:predict}e).

The prediction does not require measuring the target model's own fragilities, because they transfer between models. Replacing the target model's exponents with those of the same task and format averaged over the other models, excluding every model of its family (the four Qwen2.5 models form one family), gave a median error of 0.117--0.125, lower than the uniform decline in all 23 cells (preregistered analysis; Fig.~\ref{fig:predict}d,e), about as low as with the target model's own profile (0.110--0.118; a preregistered test that the own profile would be at least 10\% more accurate did not pass; Supplementary Table~2, R1). The transferred profile was also more accurate than three other predictions that need no damaged data of the target model: format effects alone (0.171--0.175), the intact accuracy of each measure (0.185--0.205), and the intact answer margins combined with the format effects of other models (0.133--0.142), in line with the low-dimensional structure of intact capabilities\citep{ruan2024observational}. Reading each capability's loss directly from other families at the same overall loss, without the rank-one model, did about as well (0.115--0.125; exploratory; Supplementary Note~9). At mild compression, where models keep at least 85\% of their ability, the profile still ordered the losses (Spearman 0.55--0.65; exploratory). How much a capability loses under compression is therefore largely a property of its task and format, shared across model families.

\section*{Discussion}

Across eight language models, four random perturbations and three compression methods produced strongly correlated patterns of short-answer capability loss, and how securely the task is held, how familiar the fact is and how much of the answer must be produced predicted that shared order. The regularity is not universal: quantization with GPTQ did not preserve it across models, and longer generative tasks degraded in patterns of their own.

Four preregistered tests failed, and they mark the scope of these claims (Supplementary Table~2). First, the order under GPTQ agreed with each model's own random-damage order but not between models (median correlation 0.095 over six pairs, against 0.924 for Wanda and 0.927 for SparseGPT). Second, long-form tasks (reasoning with chain of thought, code, instruction following and function calling) did not keep their position across damages: 10 of 16 task--model pairs stayed within the preregistered band, short of the required 75\%. Third, with chance set to zero for every generated measure, the shared order, its transfer to compression and the accuracy prediction held (correlations of 0.83 between damages and 0.83 with compression; 47--51\% smaller error than a uniform decline), but answers chosen from options were more robust than bare answers in five of eight models rather than the required seven. Fourth, the accuracy predictions reduced the error of a uniform decline by 44--46\% in all 23 cells but missed the preregistered absolute bound of 0.10 (0.110--0.118), an error close to the 0.117 by which the same measure differs between two runs of the same random damage at the same dose (exploratory). The order of short-answer capabilities is therefore shared across random damages and models and followed by compression within each model; the position of long-form tasks, the order under quantization across models and the absolute accuracy of single predictions are not settled by this study.

The axes account for much of the disagreement in the engineering literature. Of 13 published fragility rankings we could recompute, nine shrink, vanish or reverse once scores are corrected for chance or compared within one format, and the two that hold within one format agree with our order (Table~\ref{tab:audit}; Supplementary Table~1). Fact recall fails early in some studies and late in others because it is not one capability: secure, frequent facts survive and others do not. Obscure entities appear spared by compression\citep{wu2026asymmetric} because that study grouped facts by the popularity of the entity asked about, which has no effect of its own in our data, and did not restrict the comparison to facts the intact model knew.

\begin{table}[tbp]
\centering
\small
\caption{\textbf{Published rankings of tasks by fragility, before and after chance correction or a same-format comparison.} Numbers are from our reading of the original papers; values read from figures are approximate. ``Corrected'' means rescaled so that chance is zero. All 20 audited studies are listed in Supplementary Table~1.}
\label{tab:audit}
\renewcommand{\arraystretch}{1.18}
\begin{tabular}{@{}>{\raggedright\arraybackslash}p{2.6cm}>{\raggedright\arraybackslash}p{2.0cm}>{\raggedright\arraybackslash}p{5.0cm}>{\raggedright\arraybackslash}p{4.9cm}@{}}
\toprule
Study & Damage & Published ranking & On a common footing \\
\midrule
\multicolumn{4}{@{}l}{\textbf{Reverses}} \\[1pt]
Men et al.\citep{men2024shortgptarxiv} & layer removal & summarization (generated) collapses in the 7B models & 13B models: summarization loses 25--39\%, BoolQ 55--61\% \\
Du et al.\citep{du2025context} & longer context & Mistral-7B: GSM8K loses more raw points than MMLU ($-34.2$ vs $-20.3$) & corrected: MMLU loses 70\%, GSM8K 48\% \\
Huang et al.\citep{huang2026errors} & bit flips & code and BoolQ lose the most raw points ($-5.0$, $-4.2$) & corrected: binary multiple choice loses 12--24\%, GSM8K 3.6\% \\
Sun et al.\citep{sun2025demystifying} & hardware faults & generation ($-3.2\%$) worse than option tasks ($-1.65\%$) & on the same items, chain of thought is more robust than direct answers \\
\addlinespace[3pt]
\multicolumn{4}{@{}l}{\textbf{Shrinks or vanishes}} \\[1pt]
Shrestha et al.\citep{shrestha2026limits} & layer pruning & retention 0.81--0.91 (classification) vs 0.46--0.63 (generation) & gap roughly halves: 0.27--0.40 $\to$ 0.06--0.25 \\
Li et al.\citep{li2024evaluating} & 3-bit weights & GSM8K (generated) worse than MMLU (option likelihood) & corrected losses 42 vs 38\% (7B), 34 vs 17\% (13B), 14 vs 10\% (70B) \\
Marchisio et al.\citep{marchisio2024multilingual} & quantization & mathematics degrades most & the gap vanishes in 3 of 5 settings \\
\addlinespace[3pt]
\multicolumn{4}{@{}l}{\textbf{Holds within one format; agrees with the shared order}} \\[1pt]
Gromov et al.\citep{gromov2024unreasonable} & layer removal & GSM8K falls 50.0$\to$10.6 at 20\% removal; MMLU holds & MMLU with generated chain of thought also holds \\
Liu et al.\citep{liu2025quantreasoning} & quantization & harder mathematics fails first (same format) & holds within format \\
\bottomrule
\end{tabular}
\end{table}

The format axis also bears on how damage should be measured. Because templates fail before content, strict scoring inflates the cost of compression unevenly: near half the intact score, the strict loss of \twc{} is 2.0--2.3 times its content-only loss under all three methods and that of \jsn{} 1.0--1.8 times, whereas for bare answers the two agree (Extended Data Fig.~3). Fragility should therefore be compared within one answer format, after correcting for chance, with template failures reported separately from wrong answers. Under these rules, most published disagreements dissolve.

For deployment, the profile is a forecast: measured once, in the model itself or in other model families, it predicts the accuracy each capability keeps after compression with about half the error of a uniform decline. Because pruning and quantization remove rare answers first, as compression of image classifiers removes rare and atypical examples first\citep{hooker2019compressed}, a compressed model keeps its common knowledge and loses its long tail. In domains where the rare answer is the one that matters, such as uncommon diagnoses in medicine, this loss should be measured directly before compressed models are deployed; aggregate scores, dominated by common knowledge, understate it. This study did not, however, evaluate clinical decision-making, safety-critical deployment or the recognition of rare events in any specific domain.

The regularities along the fact and format axes are those that patients with semantic dementia show and that lesioned connectionist networks of 100 to 300 units reproduced\citep{hinton1991lesioning,rogers2004structure}. We find them in networks of 1.5 to 32 billion parameters, under random lesions and under compression. Lesions confined to particular components of language models produce deficits that depend on what was lesioned\citep{roll2026aphasias}; the regularities we report recur under all seven damages we applied, which supports the view that they follow from how distributed networks learn. The resemblance concerns what is lost and in what order, not shared mechanisms.

Damaged models give little sign of their damage (Extended Data Fig.~6): their errors are fluent answers in the requested form, their confidence falls more slowly than their accuracy (exploratory), and no label-free signal we tested predicted the loss of content across models (preregistered; leave-one-model-out $R^2$ at most 0.38). Detecting that a deployed model has been degraded therefore requires labelled checks or human oversight.

The work has limits. The models reach 32 billion parameters, and within the Qwen family the agreement between random damages declines with size (0.96, 0.91 and 0.76 from 1.5B to 14B) before recovering at 32B. Because of the failures above, our claims concern short answers. Frequencies were counted in one open corpus as a proxy for the undisclosed training data, and the fact axis was tested in two tasks whose answers are words. Why writing an answer twice is the most fragile format remains open, and a mechanistic account will need internal measurements of where knowledge is stored\citep{meng2022locating}. The comparison with patients rests on published findings, and all decoding was greedy.

Within short answers, what a damaged language model loses first is decided before the damage. Three properties of what is asked predict the order, how securely the model holds the task, how familiar the fact is and how much of the answer it must produce, and the damage only sets how fast the model moves along them. Because these axes are shared across damages and model families, they can be measured once to forecast which capabilities and facts compression will remove and, after brief calibration, how much accuracy each keeps; and because the format moves a capability as far as the task, capabilities must be compared within one answer format. For language-model research, this turns the evaluation of each new compression method from an open-ended search into a measurement along three known axes. That two regularities of semantic dementia reappear in networks of billions of parameters suggests that how robust a capability is depends on what a network has learned rather than on how it is damaged, and makes language models a system in which to study the order of forgetting.

\section*{Methods}

\subsection*{Models, tasks and damage}

\paragraph{Models.} We evaluated Qwen2.5-1.5B, 7B, 14B and 32B-Instruct\citep{qwen2025qwen25}, Llama-3.2-3B-Instruct\citep{grattafiori2024llama3,meta2024llama32}, Mistral-7B-Instruct-v0.3\citep{jiang2023mistral,mistral2024v03}, Gemma-2-9B-it\citep{gemma2024gemma2} and Phi-3.5-mini-instruct\citep{abdin2024phi3}, in bfloat16 with greedy decoding. All conditions of a model ran on the same machine and software stack, so that damaged weights are reproducible bit for bit; this was verified for every condition of the long-form extension by a fixed probe whose negative log-likelihood had to match the archived condition within 0.1\%.

\paragraph{Tasks.} Sixteen tasks were used (Extended Data Table~2). Six are generated programmatically: adding two, three or four two-digit numbers, naming a capital city (78 countries in the evaluation set), looking up a three-digit code among twelve key--code pairs, and copying a six-letter nonsense word. Ten come from SciQ\citep{welbl2017sciq}, BoolQ\citep{clark2019boolq}, ARC-Challenge\citep{clark2018arc}, CommonsenseQA\citep{talmor2019csqa}, two MMLU subsets\citep{hendrycks2021mmlu} (mathematics and physics; history, nutrition, geography and psychology), HellaSwag\citep{zellers2019hellaswag}, PIQA\citep{bisk2020piqa}, WinoGrande\citep{sakaguchi2020winogrande} and OpenBookQA\citep{mihaylov2018obqa}. Each task has 300 evaluation and 40 calibration items with 2--5 candidate answers; all items were generated once and frozen with hashes.

\paragraph{Answer formats and scoring.} Every task is posed in four formats that differ only in the final instruction. \cho{} compares the candidates by summed log-probability, each followed by the end-of-turn token, as a continuation of the \ans{} prompt. \ans{} asks for the answer alone, \jsn{} for an object \texttt{\{"answer": "..."\}}, and \twc{} for the answer written twice, separated by a comma. Generation used at most 32 new tokens. \ans{} is judged on its first non-empty line; \jsn{} and \twc{} must match their template exactly (whitespace normalized) and then contain the right content. A response is correct only if both template and content are right. The scorer also recorded, for every response, whether the template was valid and whether the correct answer was present anywhere in the output (the first standalone option letter or yes/no for tasks with listed options or yes/no answers); for other tasks, any occurrence of the gold answer counts, which favours the generated formats. The principal analyses (the shared order across damages, the format effects, the prediction of the order under compression and of accuracy) were repeated with content scoring and with chance set to zero for every generated measure (preregistered; Supplementary Table~2, V1--V2). Chance is $1/k$ for \cho{} and for generated answers to tasks with $k$ listed options or yes/no answers, and 0 otherwise. A measure is usable in a model if its rescaled intact score $(a_m(0)-\kappa_m)/(1-\kappa_m)$ is at least 0.4.

\paragraph{Untargeted damage.} Gaussian noise adds to every weight matrix in the transformer layers noise with standard deviation $r$ times that matrix's standard deviation. Random zeroing sets a random fraction $p$ of the entries of every such matrix to zero. Neuron ablation silences a random fraction $p$ of MLP hidden units by zeroing the corresponding columns of the output projection. Random layer removal removes $n$ layers chosen uniformly among all but the first and last. Random damages were evaluated with three seeds. A preregistered control removed the $n$ consecutive layers before the last one (late-layer removal; Supplementary Table~4).

\paragraph{Compression.} Wanda\citep{sun2024wanda} prunes, within each output row, the weights with the smallest product of magnitude and input-activation norm; SparseGPT\citep{frantar2023sparsegpt} prunes column-wise with inverse-Hessian weight compensation (block size 128, damping 0.01); GPTQ\citep{frantar2023gptq} quantizes column-wise with error compensation (our settings: symmetric rounding, group-wise scales, no activation reordering, damping 0.01). All three act on every linear layer of every decoder layer, sequentially by layer, using 128 segments of 512 tokens from the C4 validation set\citep{raffel2020c4} as calibration text. Sparsity ranged over $\{0.2,\dots,0.9\}$; GPTQ configurations were ordered from 4 bits with groups of 128 to 2 bits per channel. Implementations were checked against naive baselines before the main run (for Qwen2.5-7B at 50\% sparsity, perplexity rose by 30\% under Wanda against more than a hundredfold under magnitude pruning). The methods are deterministic and were run once.

\paragraph{Dose calibration.} Each model--damage cell was calibrated automatically on calibration items, with seeds separate from those of evaluation, before any damaged condition was evaluated. Calibration items share some question texts with evaluation items in two generated tasks (Supplementary Note~1). The calibration score is the mean chance-corrected score of the \ans{} measures whose intact calibration accuracy is at least 0.3 above chance. Along a fixed grid, continuous damages use the doses at which the monotonized calibration score equals 0.95, 0.85, 0.75, 0.60, 0.45, 0.30, 0.18 and 0.08; layer removal uses all layer counts up to the first that brings the score below 0.08; GPTQ keeps every configuration whose score lies in $[0.08,0.95]$, completed to at least four. In total we evaluated more than 1,100 conditions (Extended Data Table~2).

\subsection*{Fragility and comparing damages}

\paragraph{Fragility exponents.} Within each cell, equation~(\ref{eq:rank1}) is fitted by unweighted alternating least squares on $-\log c_m(d)$, using points with $0.03<c_m(d)<0.97$, averaged over seeds. A usable measure is fitted if it has at least two points in this window, and a cell is fitted if it has at least 12 fitted measures. Exponents are normalized to a geometric mean of one within each cell, so cells are compared through correlations and centred effects. For the joint fit across damages (exploratory), the same model is fitted to all seven damages of a model at once, with one exponent per measure and one strength per damage and dose, on the measures with at least four points in the window.

\paragraph{Task and format effects.} Within each cell the log exponents are decomposed as
\begin{equation}
  \log n_{t,f} = \mu + \alpha_t + \beta_f + \varepsilon_{t,f},
  \label{eq:additive}
\end{equation}
with centred task effects $\alpha_t$ and format effects $\beta_f$, fitted by least squares over tasks fitted in at least three formats.

\paragraph{Comparing damages.} Two cells are compared by the Pearson correlation of their task effects over at least eight shared tasks. The random-damage reference of a model is the mean of its task effects over the random damages in which a task was tabled (at least two); compression cells are compared with it. Seed noise was corrected by disattenuating cross-damage correlations with the seed-to-seed reliability of each damage. For the extreme-task test, the three most robust tasks and the most fragile task were removed and the correlation was compared with a null distribution obtained by shuffling task labels within a cell (1,000 permutations). For the item-level order, survival (defined below) was computed separately for each random damage and for seeds 1 and 2; for each usable measure with at least 30 items the intact model answered correctly, we correlated (Spearman) survival between two seeds of one damage and between two damages using different seeds, took the median over damages and damage pairs, and then the median over measures within a model. In Fig.~\ref{fig:order}a,c, the accuracy left in each task is its relative score in the bare-answer format, interpolated linearly at the dose at which the model's mean relative score over all fitted measures (seed-averaged) falls to one half; the few damages that never reach one half are read at their strongest setting.

\subsection*{The three axes}

\paragraph{Answer margin.} All intact-model measurements were computed on undamaged models. The answer margin of a task is the mean over items of the \cho{} log-probability of the correct candidate minus that of the best distractor. The share of the variance of damage-averaged task effects explained by the margin was computed within each model, and with each task held out of the fit. To test whether shared margins carry the agreement between models, task effects were regressed on the margin (or on intact accuracy) within each model and the residuals were correlated between models.

\paragraph{Item survival and pretraining frequency.} The survival of an item in a measure is its mean correctness over all damaged conditions of a model (all seven damages, all doses and seeds), computed for items the intact model answers correctly; for the test under each damage, it was computed from the conditions of that damage alone. For capitals, items are grouped by country, and a country is kept if the intact model answered all its items correctly. Counts were obtained with the infini-gram engine\citep{liu2024infinigram} from its index of Dolma v1.7\citep{soldaini2024dolma} (2.6 trillion tokens under the Llama-2 tokenizer) and transformed to $\log_{10}(\mathrm{count}+1)$. For capitals, the preregistered count is the co-occurrence of country and capital (the index's conjunctive query); the counts of the capital and of the country alone were used for the partial correlations. For SciQ and the key-lookup task, the count is that of the exact answer string. Partial Spearman correlations regress the ranks of both variables on the ranks of the control and correlate the residuals. Controls were the number of tokens in the answer and the intact model's \cho{} margin for the item.

\paragraph{Errors and template failures.} The error analyses are exploratory. Substitution errors were identified by matching bare answers to the 80 capital names of the item set. For yes/no questions, the share of the more common answer among bare answers containing yes or no was compared between intact models and damaged conditions with bare accuracy of at most 0.6. Failure types of \twc{} were classified from the outputs at intermediate damage. For each item failed in \twc{}, we recorded whether the same damaged model answered it correctly in \ans{}, and conversely. Function calling used single-function calls from BFCL\citep{patil2025bfcl} emitted as JSON, scored strictly and for content alone (the right function with the right arguments), in four models under two damages (Extended Data Fig.~3).

\paragraph{Intact-model confidence as an account of the format effect.} Instruction push is the summed log-probability of a measure's canonical output, followed by the end-of-turn token, under the prompt with the format instruction minus that under the question alone, averaged over items the intact model answers correctly; the preregistered predictor is the sum of the standardized push and the standardized margin within a model. The full-output format margin scores each candidate written completely in the format under the format prompt; the first-answer margin scores only the first answer and its delimiter (the answer and end token for \ans{}, the answer and closing quote for \jsn{}, the first copy and comma for \twc{}). Per-token margins are the gaps between the correct and the strongest alternative token at the decisive positions of the canonical output.

\paragraph{Separability and the axes on one scale.} The share of variance explained by equation~(\ref{eq:additive}) was computed within each cell. For the fact and format axes, survival was computed in \ans{}, \jsn{} and \twc{}, each for the units the intact model answered correctly in that format, among usable measures and in cells with at least 30 units. The frequency effect is the Spearman correlation between log answer frequency and survival in each format. For the format order, units answered correctly in all three formats were split at the median log frequency within a model and task, and the mean survival of each format was compared within each half. For the comparison of the axes on one scale (exploratory), the survival of a measure is the mean of its relative score $c_m(d)$, clipped to $[0,1]$, over all damaged conditions of a model, for measures whose intact accuracy exceeds chance by more than 0.05. Each axis was divided into four levels ordered in advance and compared by the difference between its first and last level. For the format, the levels are the four formats (\cho{} minus \twc{}, mean over tasks). For the task, the tasks were divided into quarters of four by their intact margin within each format (median over formats); ordering tasks by their own survival instead gives a larger difference (0.44), inflated by selection. For the fact, the units of capitals and SciQ in \ans{} were divided into quarters by log answer frequency (mean of the two tasks). Each difference is reported as the median over models.

\subsection*{Prediction}

\paragraph{Prediction of exponents and of facts.} For a held-out random damage, the additive fit averaged over the other three random damages predicts each measure's log exponent; for compression, the fit averaged over all four random damages does. Predictions are compared with observed log exponents by Pearson correlation and against task effects or format effects alone. Survival under random damage was pooled over the four random damages, and survival under compression was computed for each method alone, for countries (capitals) and items (SciQ, key lookup) the intact model answered correctly, in cells with at least 30 such units. The partial correlation holds the log frequency of the answer constant, or both the log frequency and the intact item margin (the choice-format score of the correct candidate minus that of the best distractor, averaged over a country's items for capitals), and is computed on ranks.

\paragraph{Prediction of accuracy.} Exponents were fitted in each random-damage cell as above, normalized within the cell and averaged in log over the random damages in which a measure was fitted (at least two); no compression data entered them. In each compression condition with a median relative score between 0.10 and 0.90, the strength was estimated by least squares, $\hat h=\sum_j n_j L_j/\sum_j n_j^2$ with $L_j=-\log c_j$, over calibration measures with $0.03<c_j<0.97$, and the accuracy of a predicted measure $t$ was $\kappa_t+(a_t(0)-\kappa_t)\exp(-n_t\hat h)$. Calibration measures were all formats of four fixed tasks, chosen before any prediction from intact data only: the 16 tasks were ranked by the median over the eight models of their within-model standardized intact margin, and the tasks ranked 2, 6, 10 and 14 were taken (WinoGrande, OpenBookQA, PIQA, copying a word). Every measure of the other twelve tasks was predicted, so each predicted task was held out with all its formats; at least three calibration measures were required. As a second scheme, each measure was predicted from all other measures of the condition, leaving it out (at least five). All exponent sources were compared on the same measures (those with both a random-damage exponent and an intact margin) and with the same calibration measures. Sources were: the model's own random-damage exponents; the transferred profile, $\exp(\bar a_{\mathrm{task}}+\bar b_{\mathrm{format}})$ with task and format effects averaged over the other models, excluding all models of the target's family (the four Qwen2.5 models form one family, each other model its own); the intact margins, a single slope relating task effects to standardized margins fitted by least squares through the origin on the other models plus their mean format effects; the intact accuracy, $\log n=g\,z$ with $z$ the within-model standardized chance-corrected intact accuracy of the measure and $g$ fitted by pooled least squares on the centred log exponents of the other models; format effects alone; and a uniform decline with every exponent one. Errors are mean absolute differences between predicted and observed accuracy over all predicted measures and informative conditions of a model and method (at least 20 points). As an exploratory reference, we computed the mean absolute difference in accuracy between two seeds of the same random damage at the same dose over the same range of doses, and compared six variants of the prediction (three ways of taking exponents and two ways of estimating the strength) by predicting each random damage from the other three; they differed by less than 0.01 and none was carried forward.

\subsection*{Other analyses, preregistration and statistics}

\paragraph{Confidence and label-free signals.} Confidence is the largest probability among the candidates after normalizing the \cho{} log-probabilities over candidates; its ability to separate correct from wrong answers was measured by the area under the receiver operating characteristic curve within each measure, with at least ten correct and ten wrong items (exploratory). For the preregistered test of label-free signals, content loss was one minus the ratio of the mean chance-corrected \cho{} and \ans{} accuracy over usable measures to its intact value. The three signals were the normalized rise in the template failure rate of \twc{} over all 16 tasks, the relative rise in negative log-likelihood on a fixed probe text, and the KL divergence from the intact model on that text. For each signal, an isotonic regression from signal to content loss was fitted on seven models and evaluated on the eighth, and $R^2$ was computed over all held-out conditions with a content loss of at most 0.5.

\paragraph{Literature audit.} For 20 studies of capability loss under damage we read the pages that report each task comparison, recorded each task's format, scoring rule, chance level and doses, and, where the numbers allowed, recomputed losses after rescaling chance to zero (Supplementary Information).

\paragraph{Preregistration.} The study proceeded in time-stamped protocols, each written with its predictions, thresholds and analysis script (locked by checksum) before the data or measurements it tests were collected. Supplementary Table~2 lists every test and its outcome. Passed: shared order across random damages within and across models; reliability; reproduction of an earlier study; additive prediction of a held-out damage; compression following the random-damage order within a model; two-effect transfer to compression; order at 32B; robustness to removing extreme tasks; format failure before content in function calling; frequency effects for capitals and SciQ, with the key-lookup control, and under each damage separately; transfer of fact survival from random damage to each compression method, beyond answer frequency; frequency effects in every format and the format order among rare and common answers; prediction of accuracy under compression from random-damage exponents, and from intact margins, against a uniform decline, also with four fixed calibration tasks, whole tasks held out and the target's family excluded; transfer of fact survival beyond both answer frequency and intact margin; the principal results under content scoring. Partial: instruction push plus margin; item-level order across damages. Undecided: stability of the measure-specific residual. Failed: across-model agreement under compression (because of GPTQ); late-layer control (too few measurable tasks); per-token margin, full-output format margin and first-answer margin as accounts of the format effect; position of long-form tasks across damages; label-free signals of content loss; an absolute error of accuracy predictions of at most 0.10; a gain of at least 10\% of the model's own profile, or of intact margins, over the profile transferred from other models; the principal results with chance set to zero for every generated measure (\cho{} more robust than \ans{} in five of eight models). Analyses not listed there are exploratory and are marked as such. The predictor in the instruction-push test came from a pilot on two models, which were excluded from its test; the first-answer margin test corrected a design flaw found after the full-output test and reused its locked analysis and thresholds. The frequency tests used item outputs that existed before they were preregistered, but no frequency had been counted; one of the two secondary counts for capitals had been seen before the SciQ test was written. The transfer test also used existing outputs; of their item-level relations, only the agreement between random damages had been computed before it was preregistered. The tests of accuracy prediction and of separability also used existing outputs: the exponents, task and format effects and their correlations with compression had been seen, but no prediction of accuracy and no frequency effect outside \ans{} had been computed. Before the intact-model test was locked, margins were standardized within each model because their scale differs between models (Supplementary Note~1). The tests of baselines, holdouts and scoring (R1--R4, V1--V2) used existing outputs after the accuracy predictions had been seen; their calibration tasks were chosen from intact data before any of their predictions was computed.

\paragraph{Statistics.} Medians over cells or pairs are reported with ranges; 95\% intervals for the main correlations come from resampling tasks with replacement (2,000 resamples). Two variants of the seed-noise correction gave 0.92--0.94.

\section*{Data availability}
Frozen item sets, per-item outputs for all conditions, pretraining counts, and all time-stamped protocol documents with their amendments (in the original Chinese, with English translations) are available at \url{https://github.com/ZhixiangWang-CN/damage-order-artifacts}.

\section*{Code availability}
Code for damage, compression, evaluation and the locked analysis scripts, with the checksums recorded in each protocol, is available in the same repository (\url{https://github.com/ZhixiangWang-CN/damage-order-artifacts}). An interactive summary of the main results is available at \url{https://zhixiangwang-cn.github.io/damage-order-artifacts/}.

\section*{Acknowledgements}
This work was supported by NIH R21-HL186024 (SPARTACUS), NIH R01-HL171376 (PASC) and NIH U01-CA268808 (HITPIRADS).

\section*{Author contributions}
Z.W. conceived the study, designed and ran the experiments, analysed the data and wrote the manuscript. U.B. supervised the project and reviewed and edited the manuscript.

\section*{Competing interests}
The authors declare no competing interests.

\bibliographystyle{unsrtnat}
\bibliography{main}

\begin{thebibliography}{51}
\providecommand{\natexlab}[1]{#1}
\providecommand{\url}[1]{\texttt{#1}}
\expandafter\ifx\csname urlstyle\endcsname\relax
  \providecommand{\doi}[1]{doi: #1}\else
  \providecommand{\doi}{doi: \begingroup \urlstyle{rm}\Url}\fi

\bibitem[Zhu et~al.(2024)Zhu, Li, Liu, Ma, and Wang]{zhu2024survey}
Xunyu Zhu, Jian Li, Yong Liu, Can Ma, and Weiping Wang.
\newblock A survey on model compression for large language models.
\newblock \emph{Transactions of the Association for Computational Linguistics}, 12:\penalty0 1556--1577, 2024.
\newblock \doi{10.1162/tacl_a_00704}.
\newblock URL \url{https://aclanthology.org/2024.tacl-1.85/}.

\bibitem[Frantar and Alistarh(2023)]{frantar2023sparsegpt}
Elias Frantar and Dan Alistarh.
\newblock {SparseGPT}: Massive language models can be accurately pruned in one-shot.
\newblock In \emph{Proceedings of the 40th International Conference on Machine Learning}, volume 202 of \emph{Proceedings of Machine Learning Research}, pages 10323--10337. PMLR, 2023.
\newblock URL \url{https://proceedings.mlr.press/v202/frantar23a.html}.

\bibitem[Sun et~al.(2024)Sun, Liu, Bair, and Kolter]{sun2024wanda}
Mingjie Sun, Zhuang Liu, Anna Bair, and J.~Zico Kolter.
\newblock A simple and effective pruning approach for large language models.
\newblock In \emph{International Conference on Learning Representations (ICLR)}, 2024.
\newblock URL \url{https://openreview.net/forum?id=PxoFut3dWW}.

\bibitem[Frantar et~al.(2023)Frantar, Ashkboos, Hoefler, and Alistarh]{frantar2023gptq}
Elias Frantar, Saleh Ashkboos, Torsten Hoefler, and Dan Alistarh.
\newblock {OPTQ}: Accurate quantization for generative pre-trained transformers.
\newblock In \emph{International Conference on Learning Representations (ICLR)}, 2023.
\newblock URL \url{https://openreview.net/forum?id=tcbBPnfwxS}.
\newblock Also known as GPTQ; arXiv:2210.17323 (``GPTQ: Accurate Post-Training Quantization for Generative Pre-trained Transformers'').

\bibitem[Huang et~al.(2026)Huang, Di, and Li]{huang2026errors}
Yafan Huang, Sheng Di, and Guanpeng Li.
\newblock Not all errors are equal: A systematic study of error propagation in large language model inference.
\newblock In \emph{Proceedings of the 40th ACM International Conference on Supercomputing (ICS '26)}, pages 972--985. ACM, 2026.
\newblock \doi{10.1145/3797905.3800513}.

\bibitem[Sun et~al.(2025)Sun, Coalson, Chen, Liu, Zhang, Hong, Fang, and Yang]{sun2025demystifying}
Yu~Sun, Zachary Coalson, Shiyang Chen, Hang Liu, Zhao Zhang, Sanghyun Hong, Bo~Fang, and Lishan Yang.
\newblock Demystifying the resilience of large language model inference: An end-to-end perspective.
\newblock In \emph{Proceedings of the International Conference for High Performance Computing, Networking, Storage and Analysis (SC '25)}, pages 1127--1144. ACM, 2025.
\newblock \doi{10.1145/3712285.3759803}.

\bibitem[Men et~al.(2024)Men, Xu, Zhang, Wang, Lin, Lu, Han, and Chen]{men2024shortgptarxiv}
Xin Men, Mingyu Xu, Qingyu Zhang, Bingning Wang, Hongyu Lin, Yaojie Lu, Xianpei Han, and Weipeng Chen.
\newblock {ShortGPT}: Layers in large language models are more redundant than you expect.
\newblock \emph{arXiv preprint arXiv:2403.03853}, 2024.
\newblock Version 3; the published version (Findings of ACL 2025) reports different summarization results.

\bibitem[Gromov et~al.(2025)Gromov, Tirumala, Shapourian, Glorioso, and Roberts]{gromov2024unreasonable}
Andrey Gromov, Kushal Tirumala, Hassan Shapourian, Paolo Glorioso, and Daniel~A. Roberts.
\newblock The unreasonable ineffectiveness of the deeper layers.
\newblock In \emph{International Conference on Learning Representations (ICLR)}, 2025.
\newblock URL \url{https://openreview.net/forum?id=ngmEcEer8a}.
\newblock arXiv:2403.17887.

\bibitem[Hooker et~al.(2019)Hooker, Courville, Clark, Dauphin, and Frome]{hooker2019compressed}
Sara Hooker, Aaron Courville, Gregory Clark, Yann Dauphin, and Andrea Frome.
\newblock What do compressed deep neural networks forget?
\newblock \emph{arXiv preprint arXiv:1911.05248}, 2019.
\newblock URL \url{https://arxiv.org/abs/1911.05248}.

\bibitem[Jaiswal et~al.(2024)Jaiswal, Gan, Du, Zhang, Wang, and Yang]{jaiswal2024compressing}
Ajay Jaiswal, Zhe Gan, Xianzhi Du, Bowen Zhang, Zhangyang Wang, and Yinfei Yang.
\newblock Compressing {LLMs}: The truth is rarely pure and never simple.
\newblock In \emph{International Conference on Learning Representations (ICLR)}, 2024.
\newblock URL \url{https://openreview.net/forum?id=B9klVS7Ddk}.

\bibitem[von Rad et~al.(2026)von Rad, Cao, and Geiger]{vonrad2026unicomp}
Jonathan von Rad, Yong Cao, and Andreas Geiger.
\newblock {UniComp}: A unified evaluation of large language model compression via pruning, quantization, and distillation.
\newblock In \emph{Proceedings of the 2026 Conference on Empirical Methods in Natural Language Processing ({EMNLP})}, pages 49567--49589, Budapest, Hungary, 2026. Association for Computational Linguistics.
\newblock URL \url{https://aclanthology.org/2026.emnlp-main.2180/}.

\bibitem[Wu et~al.(2026)Wu, Li, Semenova, and Zhong]{wu2026asymmetric}
Yuan Wu, Mairui Li, Lesia Semenova, and Chudi Zhong.
\newblock The asymmetric harms of {LLM} compression.
\newblock \emph{arXiv preprint arXiv:2608.19670}, 2026.

\bibitem[Schaeffer et~al.(2023)Schaeffer, Miranda, and Koyejo]{schaeffer2023mirage}
Rylan Schaeffer, Brando Miranda, and Sanmi Koyejo.
\newblock Are emergent abilities of large language models a mirage?
\newblock In \emph{Advances in Neural Information Processing Systems}, volume~36, pages 55565--55581, 2023.
\newblock URL \url{https://papers.nips.cc/paper_files/paper/2023/hash/adc98a266f45005c403b8311ca7e8bd7-Abstract-Conference.html}.

\bibitem[Tam et~al.(2024)Tam, Wu, Tsai, Lin, Lee, and Chen]{tam2024speak}
Zhi~Rui Tam, Cheng-Kuang Wu, Yi-Lin Tsai, Chieh-Yen Lin, Hung-yi Lee, and Yun-Nung Chen.
\newblock Let me speak freely? a study on the impact of format restrictions on large language model performance.
\newblock In \emph{Proceedings of the 2024 Conference on Empirical Methods in Natural Language Processing: Industry Track}, pages 1218--1236, Miami, Florida, US, 2024. Association for Computational Linguistics.
\newblock \doi{10.18653/v1/2024.emnlp-industry.91}.
\newblock URL \url{https://aclanthology.org/2024.emnlp-industry.91/}.

\bibitem[Wang et~al.(2024)Wang, Ma, Hu, Weber-Genzel, R{\"o}ttger, Kreuter, Hovy, and Plank]{wang2024my}
Xinpeng Wang, Bolei Ma, Chengzhi Hu, Leon Weber-Genzel, Paul R{\"o}ttger, Frauke Kreuter, Dirk Hovy, and Barbara Plank.
\newblock ``my answer is {C}'': First-token probabilities do not match text answers in instruction-tuned language models.
\newblock In \emph{Findings of the Association for Computational Linguistics: ACL 2024}, pages 7407--7416, Bangkok, Thailand, 2024. Association for Computational Linguistics.
\newblock \doi{10.18653/v1/2024.findings-acl.441}.
\newblock URL \url{https://aclanthology.org/2024.findings-acl.441/}.

\bibitem[Li et~al.(2024)Li, Ning, Wang, Liu, Shi, Yan, Dai, Yang, and Wang]{li2024evaluating}
Shiyao Li, Xuefei Ning, Luning Wang, Tengxuan Liu, Xiangsheng Shi, Shengen Yan, Guohao Dai, Huazhong Yang, and Yu~Wang.
\newblock Evaluating quantized large language models.
\newblock In \emph{Proceedings of the 41st International Conference on Machine Learning}, volume 235 of \emph{Proceedings of Machine Learning Research}, pages 28480--28524. PMLR, 2024.
\newblock URL \url{https://proceedings.mlr.press/v235/li24bb.html}.

\bibitem[Marchisio et~al.(2024)Marchisio, Dash, Chen, Aumiller, {\"U}st{\"u}n, Hooker, and Ruder]{marchisio2024multilingual}
Kelly Marchisio, Saurabh Dash, Hongyu Chen, Dennis Aumiller, Ahmet {\"U}st{\"u}n, Sara Hooker, and Sebastian Ruder.
\newblock How does quantization affect multilingual {LLMs}?
\newblock In \emph{Findings of the Association for Computational Linguistics: {EMNLP} 2024}, pages 15928--15947, Miami, Florida, USA, 2024. Association for Computational Linguistics.
\newblock \doi{10.18653/v1/2024.findings-emnlp.935}.
\newblock URL \url{https://aclanthology.org/2024.findings-emnlp.935/}.

\bibitem[Roll et~al.(2026)Roll, Kries, Gwilliams, and Shain]{roll2026aphasias}
Nathan Roll, Jill Kries, Laura Gwilliams, and Cory Shain.
\newblock Artificial aphasias in lesioned language models.
\newblock \emph{arXiv preprint arXiv:2605.16222}, 2026.

\bibitem[Rogers et~al.(2004)Rogers, Lambon~Ralph, Garrard, Bozeat, McClelland, Hodges, and Patterson]{rogers2004structure}
Timothy~T. Rogers, Matthew~A. Lambon~Ralph, Peter Garrard, Sasha Bozeat, James~L. McClelland, John~R. Hodges, and Karalyn Patterson.
\newblock Structure and deterioration of semantic memory: A neuropsychological and computational investigation.
\newblock \emph{Psychological Review}, 111\penalty0 (1):\penalty0 205--235, 2004.

\bibitem[Patterson et~al.(2007)Patterson, Nestor, and Rogers]{patterson2007where}
Karalyn Patterson, Peter~J. Nestor, and Timothy~T. Rogers.
\newblock Where do you know what you know? the representation of semantic knowledge in the human brain.
\newblock \emph{Nature Reviews Neuroscience}, 8\penalty0 (12):\penalty0 976--987, 2007.

\bibitem[Hinton and Shallice(1991)]{hinton1991lesioning}
Geoffrey~E. Hinton and Tim Shallice.
\newblock Lesioning an attractor network: Investigations of acquired dyslexia.
\newblock \emph{Psychological Review}, 98\penalty0 (1):\penalty0 74--95, 1991.

\bibitem[Cobbe et~al.(2021)Cobbe, Kosaraju, Bavarian, Chen, Jun, Kaiser, Plappert, Tworek, Hilton, Nakano, Hesse, and Schulman]{cobbe2021gsm8k}
Karl Cobbe, Vineet Kosaraju, Mohammad Bavarian, Mark Chen, Heewoo Jun, Lukasz Kaiser, Matthias Plappert, Jerry Tworek, Jacob Hilton, Reiichiro Nakano, Christopher Hesse, and John Schulman.
\newblock Training verifiers to solve math word problems.
\newblock \emph{arXiv preprint arXiv:2110.14168}, 2021.

\bibitem[Chen et~al.(2021)]{chen2021humaneval}
Mark Chen et~al.
\newblock Evaluating large language models trained on code.
\newblock \emph{arXiv preprint arXiv:2107.03374}, 2021.

\bibitem[Austin et~al.(2021)]{austin2021mbpp}
Jacob Austin et~al.
\newblock Program synthesis with large language models.
\newblock \emph{arXiv preprint arXiv:2108.07732}, 2021.

\bibitem[Zhou et~al.(2023)Zhou, Lu, Mishra, Brahma, Basu, Luan, Zhou, and Hou]{zhou2023ifeval}
Jeffrey Zhou, Tianjian Lu, Swaroop Mishra, Siddhartha Brahma, Sujoy Basu, Yi~Luan, Denny Zhou, and Le~Hou.
\newblock Instruction-following evaluation for large language models.
\newblock \emph{arXiv preprint arXiv:2311.07911}, 2023.

\bibitem[Patil et~al.(2025)Patil, Mao, Yan, Ji, Suresh, Stoica, and Gonzalez]{patil2025bfcl}
Shishir~G. Patil, Huanzhi Mao, Fanjia Yan, Charlie Cheng-Jie Ji, Vishnu Suresh, Ion Stoica, and Joseph~E. Gonzalez.
\newblock The {Berkeley} function calling leaderboard ({BFCL}): From tool use to agentic evaluation of large language models.
\newblock In \emph{International Conference on Machine Learning}, 2025.

\bibitem[Kandpal et~al.(2023)Kandpal, Deng, Roberts, Wallace, and Raffel]{kandpal2023large}
Nikhil Kandpal, Haikang Deng, Adam Roberts, Eric Wallace, and Colin Raffel.
\newblock Large language models struggle to learn long-tail knowledge.
\newblock In \emph{Proceedings of the 40th International Conference on Machine Learning}, volume 202 of \emph{Proceedings of Machine Learning Research}, pages 15696--15707. PMLR, 2023.
\newblock URL \url{https://proceedings.mlr.press/v202/kandpal23a.html}.

\bibitem[Soldaini et~al.(2024)Soldaini, Kinney, Bhagia, Schwenk, Atkinson, Authur, Bogin, Chandu, Dumas, Elazar, et~al.]{soldaini2024dolma}
Luca Soldaini, Rodney Kinney, Akshita Bhagia, Dustin Schwenk, David Atkinson, Russell Authur, Ben Bogin, Khyathi Chandu, Jennifer Dumas, Yanai Elazar, et~al.
\newblock Dolma: an open corpus of three trillion tokens for language model pretraining research.
\newblock In \emph{Proceedings of the 62nd Annual Meeting of the Association for Computational Linguistics}, 2024.

\bibitem[Liu et~al.(2024)Liu, Min, Zettlemoyer, Choi, and Hajishirzi]{liu2024infinigram}
Jiacheng Liu, Sewon Min, Luke Zettlemoyer, Yejin Choi, and Hannaneh Hajishirzi.
\newblock Infini-gram: Scaling unbounded n-gram language models to a trillion tokens.
\newblock In \emph{First Conference on Language Modeling}, 2024.

\bibitem[Ruan et~al.(2024)Ruan, Maddison, and Hashimoto]{ruan2024observational}
Yangjun Ruan, Chris~J. Maddison, and Tatsunori Hashimoto.
\newblock Observational scaling laws and the predictability of language model performance.
\newblock In A.~Globerson, L.~Mackey, D.~Belgrave, A.~Fan, U.~Paquet, J.~Tomczak, and C.~Zhang, editors, \emph{Advances in Neural Information Processing Systems}, volume~37, pages 15841--15892. Curran Associates, Inc., 2024.
\newblock URL \url{https://papers.nips.cc/paper_files/paper/2024/hash/1cded4f97cf5f01a284c574110b7e3b9-Abstract-Conference.html}.

\bibitem[Du et~al.(2025)Du, Tian, Ronanki, Rongali, Bodapati, Galstyan, Wells, Schwartz, Huerta, and Peng]{du2025context}
Yufeng Du, Minyang Tian, Srikanth Ronanki, Subendhu Rongali, Sravan~Babu Bodapati, Aram Galstyan, Azton Wells, Roy Schwartz, Eliu~A. Huerta, and Hao Peng.
\newblock Context length alone hurts {LLM} performance despite perfect retrieval.
\newblock In \emph{Findings of the Association for Computational Linguistics: {EMNLP} 2025}, pages 23281--23298, Suzhou, China, 2025. Association for Computational Linguistics.
\newblock \doi{10.18653/v1/2025.findings-emnlp.1264}.
\newblock URL \url{https://aclanthology.org/2025.findings-emnlp.1264/}.

\bibitem[Shrestha et~al.(2026)Shrestha, Shrestha, Nepal, Kim, and Ross]{shrestha2026limits}
Safal Shrestha, Anubhav Shrestha, Aadim Nepal, Minwu Kim, and Keith Ross.
\newblock On the limits of layer pruning for generative reasoning in large language models.
\newblock \emph{arXiv preprint arXiv:2602.01997}, 2026.
\newblock Version 4.

\bibitem[Liu et~al.(2025)Liu, Sun, Zhang, Bai, Yu, Yu, Yuan, and Hou]{liu2025quantreasoning}
Ruikang Liu, Yuxuan Sun, Manyi Zhang, Haoli Bai, Xianzhi Yu, Tiezheng Yu, Chun Yuan, and Lu~Hou.
\newblock Quantization hurts reasoning? {An} empirical study on quantized reasoning models.
\newblock In \emph{Conference on Language Modeling (COLM)}, 2025.
\newblock URL \url{https://openreview.net/forum?id=BM192Ps5Nv}.
\newblock arXiv:2504.04823.

\bibitem[Meng et~al.(2022)Meng, Bau, Andonian, and Belinkov]{meng2022locating}
Kevin Meng, David Bau, Alex Andonian, and Yonatan Belinkov.
\newblock Locating and editing factual associations in {GPT}.
\newblock In \emph{Advances in Neural Information Processing Systems}, volume~35, pages 17359--17372, 2022.
\newblock URL \url{https://papers.nips.cc/paper_files/paper/2022/hash/6f1d43d5a82a37e89b0665b33bf3a182-Abstract-Conference.html}.

\bibitem[{Qwen Team} et~al.(2024){Qwen Team}, Yang, Yang, Zhang, Hui, Zheng, Yu, Li, Liu, Huang, Wei, Lin, Yang, Tu, Zhang, Yang, Yang, Zhou, Lin, Dang, Lu, Bao, Yang, Yu, Li, Xue, Zhang, Zhu, Men, Lin, Li, Tang, Xia, Ren, Ren, Fan, Su, Zhang, Wan, Liu, Cui, Zhang, and Qiu]{qwen2025qwen25}
{Qwen Team}, An~Yang, Baosong Yang, Beichen Zhang, Binyuan Hui, Bo~Zheng, Bowen Yu, Chengyuan Li, Dayiheng Liu, Fei Huang, Haoran Wei, Huan Lin, Jian Yang, Jianhong Tu, Jianwei Zhang, Jianxin Yang, Jiaxi Yang, Jingren Zhou, Junyang Lin, Kai Dang, Keming Lu, Keqin Bao, Kexin Yang, Le~Yu, Mei Li, Mingfeng Xue, Pei Zhang, Qin Zhu, Rui Men, Runji Lin, Tianhao Li, Tianyi Tang, Tingyu Xia, Xingzhang Ren, Xuancheng Ren, Yang Fan, Yang Su, Yichang Zhang, Yu~Wan, Yuqiong Liu, Zeyu Cui, Zhenru Zhang, and Zihan Qiu.
\newblock {Qwen2.5} technical report.
\newblock \emph{arXiv preprint arXiv:2412.15115}, 2024.

\bibitem[Grattafiori et~al.(2024)]{grattafiori2024llama3}
Aaron Grattafiori et~al.
\newblock The {Llama} 3 herd of models.
\newblock \emph{arXiv preprint arXiv:2407.21783}, 2024.

\bibitem[{Meta AI}(2024)]{meta2024llama32}
{Meta AI}.
\newblock Llama 3.2: Revolutionizing edge {AI} and vision with open, customizable models.
\newblock \url{https://ai.meta.com/blog/llama-3-2-connect-2024-vision-edge-mobile-devices/}, 2024.
\newblock Blog post, 25 September 2024.

\bibitem[Jiang et~al.(2023)Jiang, Sablayrolles, Mensch, Bamford, Chaplot, de~las Casas, Bressand, Lengyel, Lample, Saulnier, Lavaud, Lachaux, Stock, Le~Scao, Lavril, Wang, Lacroix, and El~Sayed]{jiang2023mistral}
Albert~Q. Jiang, Alexandre Sablayrolles, Arthur Mensch, Chris Bamford, Devendra~Singh Chaplot, Diego de~las Casas, Florian Bressand, Gianna Lengyel, Guillaume Lample, Lucile Saulnier, L{\'e}lio~Renard Lavaud, Marie-Anne Lachaux, Pierre Stock, Teven Le~Scao, Thibaut Lavril, Thomas Wang, Timoth{\'e}e Lacroix, and William El~Sayed.
\newblock {Mistral} {7B}.
\newblock \emph{arXiv preprint arXiv:2310.06825}, 2023.

\bibitem[{Mistral AI}(2024)]{mistral2024v03}
{Mistral AI}.
\newblock {Mistral-7B-Instruct-v0.3} model card.
\newblock \url{https://huggingface.co/mistralai/Mistral-7B-Instruct-v0.3}, 2024.
\newblock Released 22 May 2024; accessed 29 October 2026.

\bibitem[{Gemma Team} et~al.(2024){Gemma Team}, Riviere, et~al.]{gemma2024gemma2}
{Gemma Team}, Morgane Riviere, et~al.
\newblock {Gemma} 2: Improving open language models at a practical size.
\newblock \emph{arXiv preprint arXiv:2408.00118}, 2024.

\bibitem[Abdin et~al.(2024)]{abdin2024phi3}
Marah Abdin et~al.
\newblock {Phi-3} technical report: A highly capable language model locally on your phone.
\newblock \emph{arXiv preprint arXiv:2404.14219}, 2024.

\bibitem[Welbl et~al.(2017)Welbl, Liu, and Gardner]{welbl2017sciq}
Johannes Welbl, Nelson~F. Liu, and Matt Gardner.
\newblock Crowdsourcing multiple choice science questions.
\newblock In \emph{Proceedings of the 3rd Workshop on Noisy User-generated Text}, pages 94--106, Copenhagen, Denmark, 2017. Association for Computational Linguistics.
\newblock \doi{10.18653/v1/W17-4413}.
\newblock URL \url{https://aclanthology.org/W17-4413/}.

\bibitem[Clark et~al.(2019)Clark, Lee, Chang, Kwiatkowski, Collins, and Toutanova]{clark2019boolq}
Christopher Clark, Kenton Lee, Ming-Wei Chang, Tom Kwiatkowski, Michael Collins, and Kristina Toutanova.
\newblock {BoolQ}: Exploring the surprising difficulty of natural yes/no questions.
\newblock In \emph{Proceedings of the 2019 Conference of the North {A}merican Chapter of the Association for Computational Linguistics: Human Language Technologies, Volume 1 (Long and Short Papers)}, pages 2924--2936, Minneapolis, Minnesota, 2019. Association for Computational Linguistics.
\newblock \doi{10.18653/v1/N19-1300}.
\newblock URL \url{https://aclanthology.org/N19-1300/}.

\bibitem[Clark et~al.(2018)Clark, Cowhey, Etzioni, Khot, Sabharwal, Schoenick, and Tafjord]{clark2018arc}
Peter Clark, Isaac Cowhey, Oren Etzioni, Tushar Khot, Ashish Sabharwal, Carissa Schoenick, and Oyvind Tafjord.
\newblock Think you have solved question answering? {Try} {ARC}, the {AI2} reasoning challenge.
\newblock \emph{arXiv preprint arXiv:1803.05457}, 2018.

\bibitem[Talmor et~al.(2019)Talmor, Herzig, Lourie, and Berant]{talmor2019csqa}
Alon Talmor, Jonathan Herzig, Nicholas Lourie, and Jonathan Berant.
\newblock {CommonsenseQA}: A question answering challenge targeting commonsense knowledge.
\newblock In \emph{Proceedings of the 2019 Conference of the North {A}merican Chapter of the Association for Computational Linguistics: Human Language Technologies, Volume 1 (Long and Short Papers)}, pages 4149--4158, Minneapolis, Minnesota, 2019. Association for Computational Linguistics.
\newblock \doi{10.18653/v1/N19-1421}.
\newblock URL \url{https://aclanthology.org/N19-1421/}.

\bibitem[Hendrycks et~al.(2021)Hendrycks, Burns, Basart, Zou, Mazeika, Song, and Steinhardt]{hendrycks2021mmlu}
Dan Hendrycks, Collin Burns, Steven Basart, Andy Zou, Mantas Mazeika, Dawn Song, and Jacob Steinhardt.
\newblock Measuring massive multitask language understanding.
\newblock In \emph{International Conference on Learning Representations (ICLR)}, 2021.
\newblock URL \url{https://openreview.net/forum?id=d7KBjmI3GmQ}.

\bibitem[Zellers et~al.(2019)Zellers, Holtzman, Bisk, Farhadi, and Choi]{zellers2019hellaswag}
Rowan Zellers, Ari Holtzman, Yonatan Bisk, Ali Farhadi, and Yejin Choi.
\newblock {HellaSwag}: Can a machine really finish your sentence?
\newblock In \emph{Proceedings of the 57th Annual Meeting of the Association for Computational Linguistics}, pages 4791--4800, Florence, Italy, 2019. Association for Computational Linguistics.
\newblock \doi{10.18653/v1/P19-1472}.
\newblock URL \url{https://aclanthology.org/P19-1472/}.

\bibitem[Bisk et~al.(2020)Bisk, Zellers, Le~Bras, Gao, and Choi]{bisk2020piqa}
Yonatan Bisk, Rowan Zellers, Ronan Le~Bras, Jianfeng Gao, and Yejin Choi.
\newblock {PIQA}: Reasoning about physical commonsense in natural language.
\newblock In \emph{Proceedings of the AAAI Conference on Artificial Intelligence}, volume~34, pages 7432--7439, 2020.
\newblock \doi{10.1609/aaai.v34i05.6239}.

\bibitem[Sakaguchi et~al.(2020)Sakaguchi, Le~Bras, Bhagavatula, and Choi]{sakaguchi2020winogrande}
Keisuke Sakaguchi, Ronan Le~Bras, Chandra Bhagavatula, and Yejin Choi.
\newblock {WinoGrande}: An adversarial {Winograd} schema challenge at scale.
\newblock In \emph{Proceedings of the AAAI Conference on Artificial Intelligence}, volume~34, pages 8732--8740, 2020.
\newblock \doi{10.1609/aaai.v34i05.6399}.

\bibitem[Mihaylov et~al.(2018)Mihaylov, Clark, Khot, and Sabharwal]{mihaylov2018obqa}
Todor Mihaylov, Peter Clark, Tushar Khot, and Ashish Sabharwal.
\newblock Can a suit of armor conduct electricity? {A} new dataset for open book question answering.
\newblock In \emph{Proceedings of the 2018 Conference on Empirical Methods in Natural Language Processing}, pages 2381--2391, Brussels, Belgium, 2018. Association for Computational Linguistics.
\newblock \doi{10.18653/v1/D18-1260}.
\newblock URL \url{https://aclanthology.org/D18-1260/}.

\bibitem[Raffel et~al.(2020)Raffel, Shazeer, Roberts, Lee, Narang, Matena, Zhou, Li, and Liu]{raffel2020c4}
Colin Raffel, Noam Shazeer, Adam Roberts, Katherine Lee, Sharan Narang, Michael Matena, Yanqi Zhou, Wei Li, and Peter~J. Liu.
\newblock Exploring the limits of transfer learning with a unified text-to-text transformer.
\newblock \emph{Journal of Machine Learning Research}, 21\penalty0 (140):\penalty0 1--67, 2020.

\end{thebibliography}

\clearpage
\section*{Extended Data}
\setcounter{table}{0}
\renewcommand{\tablename}{Extended Data Table}
\begin{table}[htbp]
\centering\footnotesize
\caption{\textbf{Fit of the rank-one model.} For each model and damage, the number of measures fitted (of 64) and the share of the variance of $-\log c_m(d)$ explained by equation~(\ref{eq:rank1}) within the fitting window. Last column, the joint fit of all seven damages with one exponent per measure (exploratory), and in brackets the median over damages of the separate fits. --, not fitted (fewer than 12 measures or fewer than three doses).}
\label{tab:ed_fit}
\setlength{\tabcolsep}{3.5pt}
\begin{tabular}{@{}lcccccccc@{}}
\toprule
Model & Noise & Zeroing & Neuron & Layers & Wanda & SparseGPT & GPTQ & Joint (separate) \\
\midrule
Qwen2.5-1.5B & 41, 0.95 & 41, 0.94 & 41, 0.94 & 33, 0.84 & 39, 0.85 & 38, 0.78 & 32, 0.85 & 0.82 (0.85) \\
Qwen2.5-7B & 54, 0.94 & 55, 0.92 & 56, 0.93 & 52, 0.86 & 47, 0.90 & 50, 0.89 & 43, 0.76 & 0.78 (0.90) \\
Qwen2.5-14B & 59, 0.92 & 61, 0.87 & 61, 0.89 & 61, 0.74 & 55, 0.88 & 55, 0.84 & 54, 0.70 & 0.74 (0.87) \\
Qwen2.5-32B & 61, 0.91 & 62, 0.91 & 63, 0.91 & 64, 0.81 & 58, 0.91 & 56, 0.90 & 46, 0.91 & 0.74 (0.91) \\
Llama-3.2-3B & 44, 0.90 & 44, 0.88 & 45, 0.90 & 41, 0.79 & 43, 0.84 & 40, 0.77 & 37, 0.77 & 0.72 (0.84) \\
Mistral-7B & 43, 0.93 & 44, 0.89 & 45, 0.92 & 43, 0.77 & 39, 0.83 & 38, 0.90 & 21, 0.93 & 0.77 (0.90) \\
Gemma-2-9B & 49, 0.90 & 52, 0.88 & 54, 0.87 & 53, 0.84 & 43, 0.92 & 46, 0.87 & 37, 0.83 & 0.77 (0.87) \\
Phi-3.5-mini & 41, 0.92 & 39, 0.88 & 46, 0.88 & 38, 0.88 & 42, 0.80 & 42, 0.82 & 38, 0.71 & 0.72 (0.88) \\
\bottomrule
\end{tabular}
\end{table}

\begin{table}[htbp]
\centering\footnotesize
\caption{\textbf{Tasks and calibrated doses.} \textbf{a}, The 16 tasks: source, what the bare answer is, number of candidates in \cho{} and number of evaluation items. Chance in \ans{}, \jsn{} and \twc{} is $1/k$ for tasks with listed options or yes/no answers and 0 otherwise. \textbf{b}, Calibrated doses: range over models of the weakest and strongest dose, and number of conditions per model in the main analyses (doses $\times$ seeds; random damages with three seeds); with controls and extensions, more than 1,100 conditions were evaluated. Noise, standard deviation relative to each matrix; zeroing and neuron ablation, fraction removed; layers, number removed; Wanda and SparseGPT, sparsity; GPTQ, index of the configuration, ordered from 4 bits with groups of 128 to 2 bits per channel.}
\label{tab:ed_tasks}
{\raggedright\textbf{a}\par}\smallskip
\begin{tabular}{@{}ll>{\raggedright\arraybackslash}p{4.2cm}>{\raggedright\arraybackslash}p{2.6cm}cc@{}}
\toprule
Task & Code & Source & Bare answer & $k$ & Items \\
\midrule
Add 2 & ARITH1 & generated & number & 4 & 300 \\
Add 3 & ARITH2 & generated & number & 4 & 300 \\
Add 4 & ARITH3 & generated & number & 4 & 300 \\
Capitals & FACT & generated (78 countries) & city name & 4 & 300 \\
Key lookup & RETR & generated & three-digit code & 4 & 300 \\
Copy word & COPY & generated & nonsense word & 4 & 300 \\
SciQ & SCIQ & SciQ & word or phrase & 4 & 300 \\
BoolQ & BOOLQ & BoolQ & yes or no & 2 & 300 \\
ARC-C & ARC & ARC-Challenge & option letter & 4 & 300 \\
CSQA & CSQA & CommonsenseQA & option letter & 5 & 300 \\
MMLU math & MMLU\_CALC & MMLU (mathematics, physics) & option letter & 4 & 300 \\
MMLU recall & MMLU\_MEM & MMLU (history, nutrition, geography, psychology) & option letter & 4 & 300 \\
HellaSwag & HELLASWAG & HellaSwag & option letter & 4 & 300 \\
PIQA & PIQA & PIQA & option letter & 2 & 300 \\
WinoGrande & WINOGRANDE & WinoGrande & option letter & 2 & 300 \\
OpenBookQA & OBQA & OpenBookQA & option letter & 4 & 300 \\
\bottomrule
\end{tabular}
\par\bigskip{\raggedright\textbf{b}\par}\smallskip
\begin{tabular}{@{}lccccccc@{}}
\toprule
Model & Noise & Zeroing & Neuron & Layers & Wanda & SparseGPT & GPTQ \\
\midrule
Qwen2.5-1.5B & 24 & 24 & 24 & 12 & 8 & 8 & 4 \\
Qwen2.5-7B & 24 & 24 & 24 & 9 & 8 & 8 & 4 \\
Qwen2.5-14B & 24 & 24 & 24 & 21 & 8 & 8 & 4 \\
Qwen2.5-32B & 24 & 24 & 24 & 30 & 8 & 8 & 4 \\
Llama-3.2-3B & 24 & 24 & 24 & 18 & 8 & 8 & 4 \\
Mistral-7B & 24 & 24 & 24 & 18 & 8 & 8 & 3 \\
Gemma-2-9B & 24 & 24 & 24 & 27 & 8 & 8 & 4 \\
Phi-3.5-mini & 24 & 24 & 24 & 9 & 8 & 8 & 4 \\
\midrule
Dose range & 0.018--0.46 & 0.0046--0.24 & 0.0084--0.53 & 1--16 & 0.17--0.75 & 0.19--0.83 & 1--5 \\
\bottomrule
\end{tabular}
\end{table}

\clearpage
\setcounter{figure}{0}
\renewcommand{\figurename}{Extended Data Fig.}
\begin{figure}[p]
\centering
\includegraphics[width=\linewidth,height=0.62\textheight,keepaspectratio]{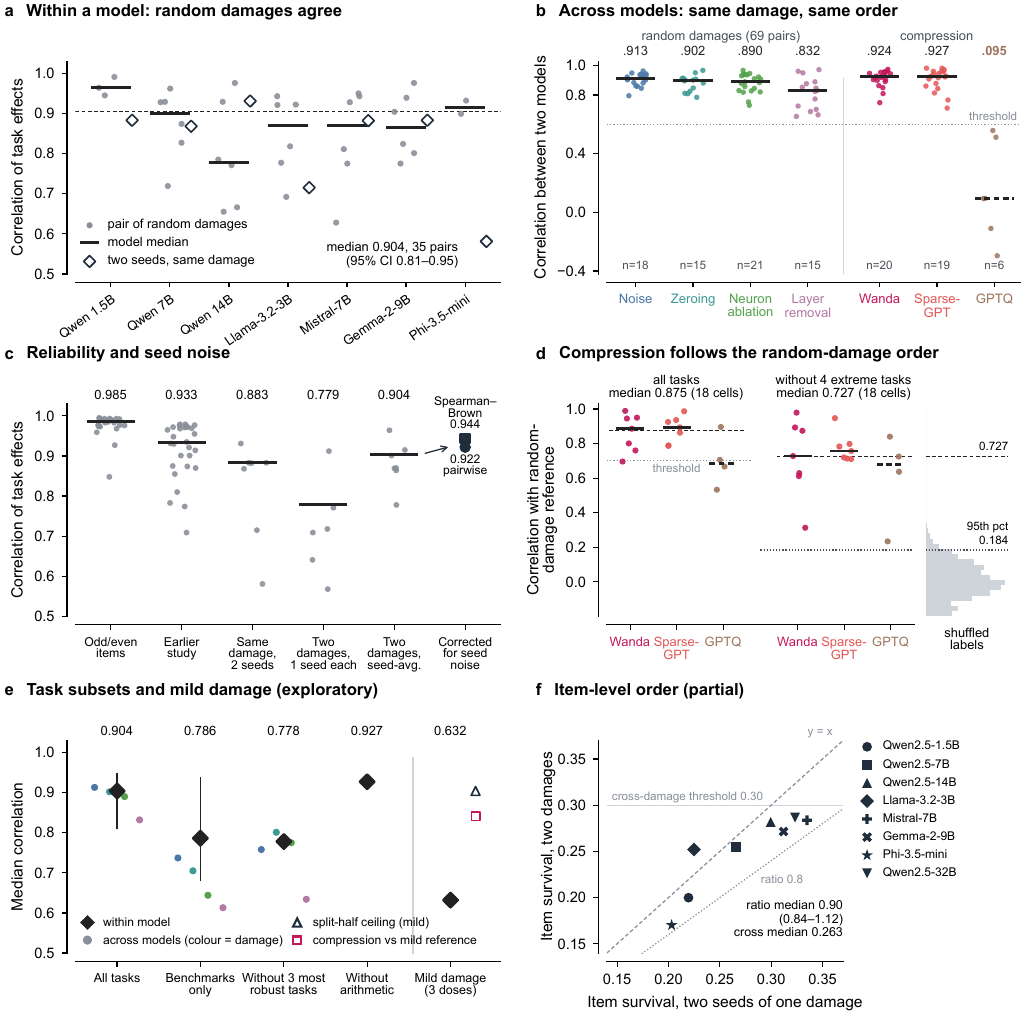}
\caption{\textbf{Agreement in detail.} Points are Pearson correlations of task effects between two cells; bars, medians; colours, damages. \textbf{a}, Between two random damages within a model (35 pairs; median 0.904, 95\% interval 0.81--0.95; every model $\ge0.778$; preregistered); diamonds, two seeds of one damage. \textbf{b}, Between models under the same damage (preregistered): random damages 0.832--0.913 (69 pairs); Wanda 0.924, SparseGPT 0.927, GPTQ 0.095 (six pairs; the test fails because of GPTQ). Dotted line, threshold. \textbf{c}, Reliability: odd versus even items (0.985), reproduction of an earlier batch (0.933), two seeds (0.883), two damages from single seeds (0.779) and from seed-averaged effects (0.904); right, agreement corrected for seed noise (0.922 and 0.944; exploratory). \textbf{d}, Compression against the same model's random-damage reference (18 cells, median 0.875; preregistered); right, without the three most robust tasks and the most fragile task (0.727) against shuffled task labels (grey, 1,000 permutations; 95th percentile 0.184). \textbf{e}, Task subsets and mild damage (first three doses; exploratory): within-model (diamonds) and across-model (points) agreement; bars, 95\% intervals. \textbf{f}, Item-level order (preregistered, partial): agreement of item survival between two damages against two seeds of one damage, per model; ratio 0.90 (0.84--1.12), both agreements 0.17--0.33.}
\label{fig:ed1}
\end{figure}
\clearpage
\begin{figure}[p]
\centering
\includegraphics[width=\linewidth,height=0.62\textheight,keepaspectratio]{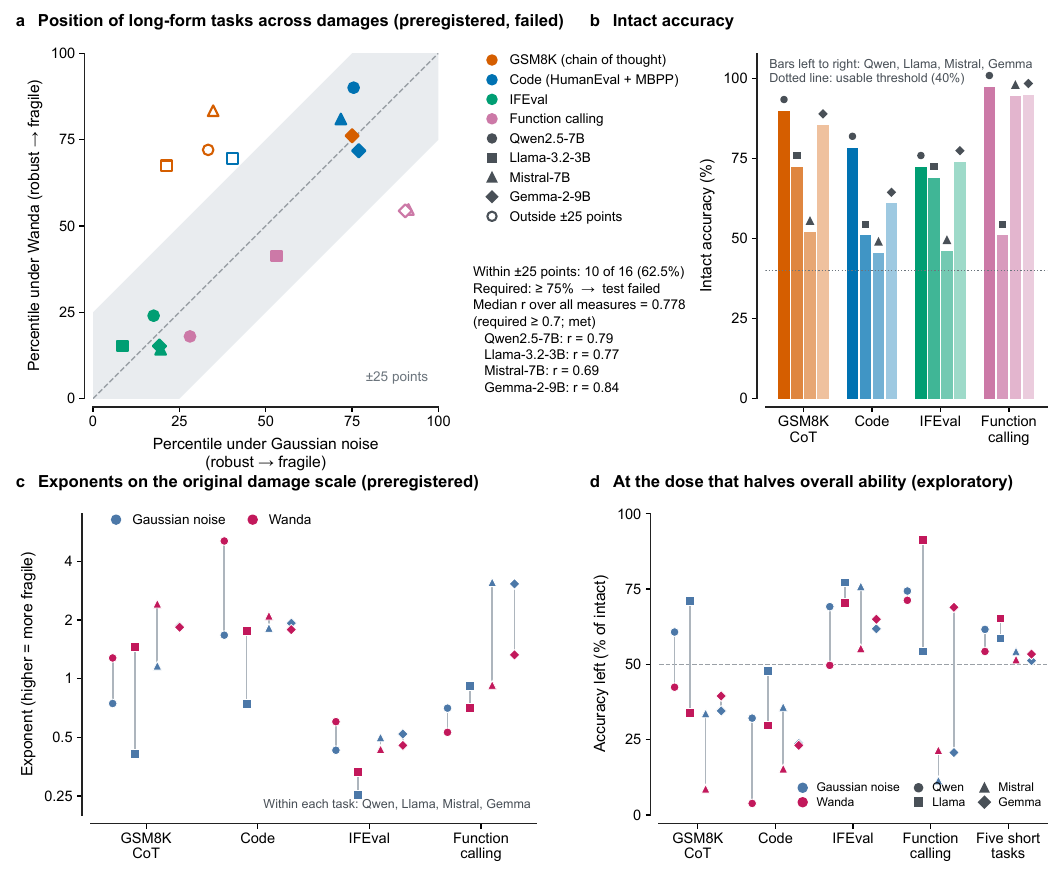}
\caption{\textbf{Long-form tasks do not keep their position across damages (preregistered, failed).} GSM8K with chain of thought, code (HumanEval and MBPP, unit tests), IFEval and function calling (BFCL), 200 items each, in four models under Gaussian noise and Wanda pruning. Colours, tasks; shapes, models. \textbf{a}, Percentile position of each task among all measures of its model (0, most robust) under noise ($x$) and Wanda ($y$); grey band, $\pm25$ points; open symbols, pairs outside the band. Ten of 16 pairs (62.5\%) lie within the band, short of the required 75\%; the median correlation of log exponents between the two damages (0.778) met its criterion. GSM8K became more fragile under Wanda in three models; IFEval stayed between the 9th and 24th percentile everywhere. \textbf{b}, Intact accuracy; dotted line, the 40\% usability threshold. \textbf{c}, Exponent of each task under noise and Wanda (log scale; lines join one model). \textbf{d}, Accuracy left at the dose that halves each model's overall ability, with the mean of five short-answer tasks for reference (exploratory).}
\label{fig:ed2}
\end{figure}
\clearpage
\begin{figure}[p]
\centering
\includegraphics[width=\linewidth,height=0.62\textheight,keepaspectratio]{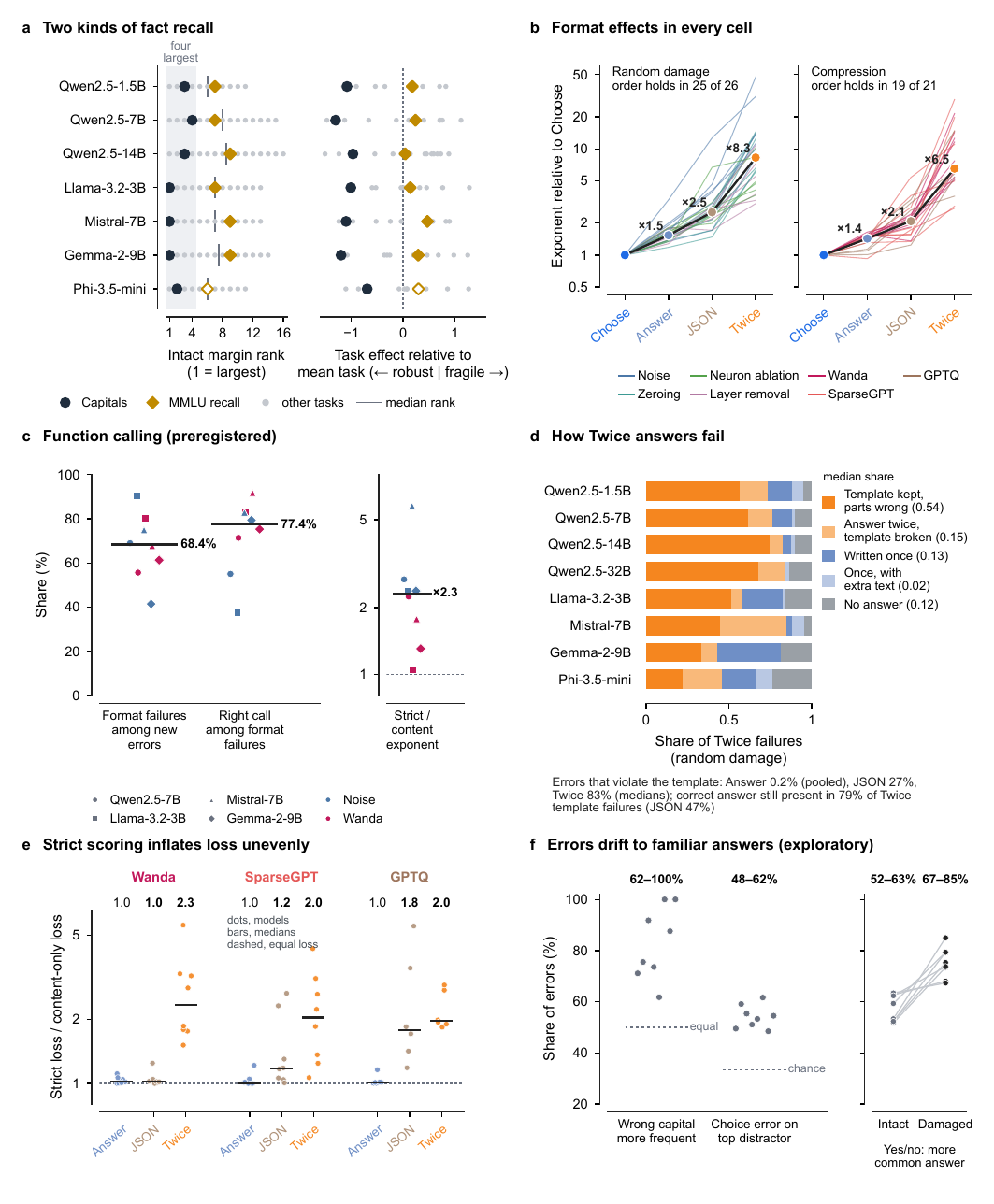}
\caption{\textbf{The task and format axes in detail.} \textbf{a}, Two kinds of fact recall: rank of each task's intact margin (left; 1, largest) and damage-averaged task effect (right; positive, more fragile) in each model. Capitals (dark circles) have one of the four largest margins and are among the most robust tasks in every model; MMLU knowledge questions (diamonds) have median margins and are more fragile than the average task. \textbf{b}, Fragility of each format relative to \cho{} in every cell (lines; log scale). Medians 1.5, 2.5 and 8.3 under random damage (26 cells; order holds in 25) and 1.4, 2.1 and 6.5 under compression (21 cells; 19). \textbf{c}, Function calling (preregistered; four models, noise and Wanda): share of damage-induced errors that are format failures (68.4\%), share of these with the right function and arguments (77.4\%), and strict over content-only exponent (2.3). \textbf{d}, How \twc{} answers fail at intermediate damage (exploratory): template kept with wrong parts 0.54, answer present twice but template broken 0.15, answer written once 0.13, no answer 0.12; 83\% of \twc{} errors violate the template, 79\% of these contain the correct answer. \textbf{e}, Strict over content-only loss for each format under each compression method (exploratory). \textbf{f}, Errors drift towards familiar answers (exploratory): wrong capitals more frequent than the correct one (62--100\%); choice errors on the most frequent distractor (48--62\%; chance 33\%); more common yes/no answer in intact and damaged models.}
\label{fig:ed3}
\end{figure}
\clearpage
\begin{figure}[p]
\centering
\includegraphics[width=\linewidth,height=0.62\textheight,keepaspectratio]{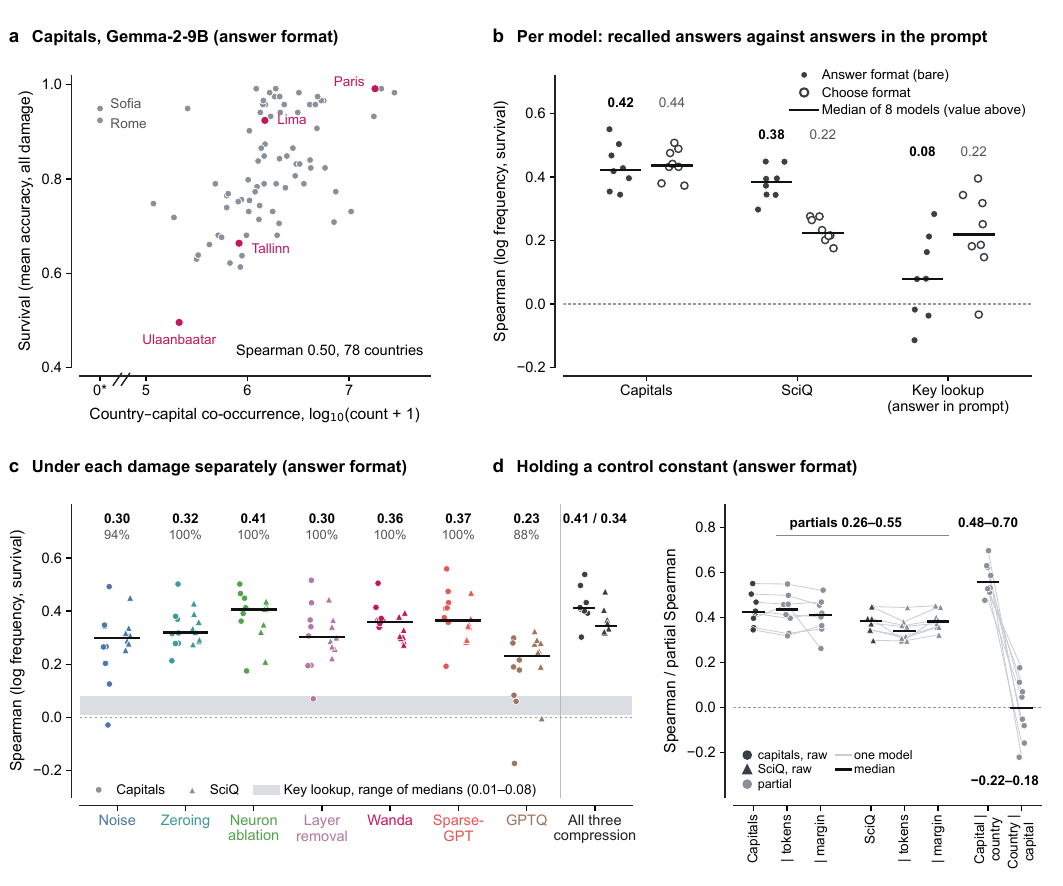}
\caption{\textbf{The fact axis per model: rare answers fail first.} Survival, mean correctness over all damaged conditions for items the intact model answered correctly; frequency, $\log_{10}(\mathrm{count}+1)$ in Dolma. Dots, models; bars, medians. \textbf{a}, Survival of each capital against country--capital co-occurrence in Gemma-2-9B (Spearman 0.50). *, Bulgaria--Sofia and Italy--Rome returned a count of 0 under the default approximation of the index and are drawn left of the axis break; re-queried, they have 541{,}362 and 896{,}384 co-occurrences, and the capitals results become stronger (Supplementary Note~4). \textbf{b}, Spearman correlation of frequency with survival for capitals (median 0.42), SciQ (0.38) and the key-lookup control (0.08); filled, \ans{} (preregistered); open, \cho{} (exploratory). \textbf{c}, The same under each damage alone and under the three compression methods pooled (preregistered); grey band, key-lookup control (0.01--0.08). \textbf{d}, Partial correlations holding answer length or the item margin constant (0.26--0.55), and for capitals, capital frequency with the country held constant (0.48--0.70) and the reverse ($-0.22$ to $0.18$).}
\label{fig:ed4}
\end{figure}
\clearpage
\begin{figure}[p]
\centering
\includegraphics[width=\linewidth,height=0.62\textheight,keepaspectratio]{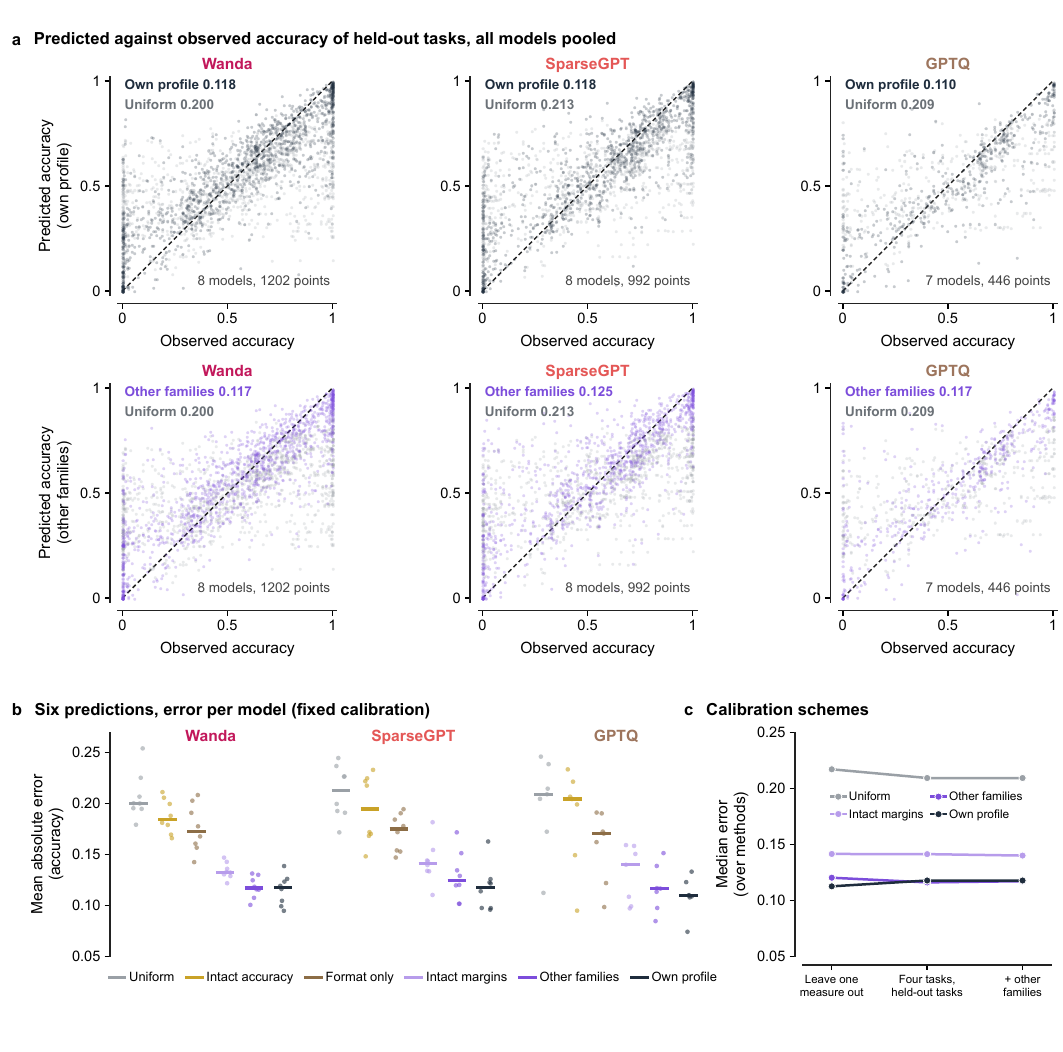}
\caption{\textbf{Prediction of accuracy under compression, every model.} \textbf{a}, Predicted against observed accuracy of every measure of the twelve tasks held out from calibration, in every informative compression condition, all models pooled, for each method; the strength of each condition was estimated on four fixed calibration tasks. Top, the model's own random-damage profile (dark); bottom, the profile transferred from other model families (purple); grey, uniform decline. Numbers, median absolute error over models. \textbf{b}, Error of each model for six predictions under the same calibration: uniform decline, intact accuracy, format effects alone, intact margins, the profile transferred from other model families, and the model's own profile; bars, medians. Predictions that borrow from other models exclude the target model's family. \textbf{c}, Median error under three calibration schemes: leaving out one measure at a time, four fixed calibration tasks with whole tasks held out, and the same excluding the target's family for predictions that borrow from other models. All panels from preregistered analyses.}
\label{fig:ed5}
\end{figure}
\clearpage
\begin{figure}[p]
\centering
\includegraphics[width=\linewidth,height=0.62\textheight,keepaspectratio]{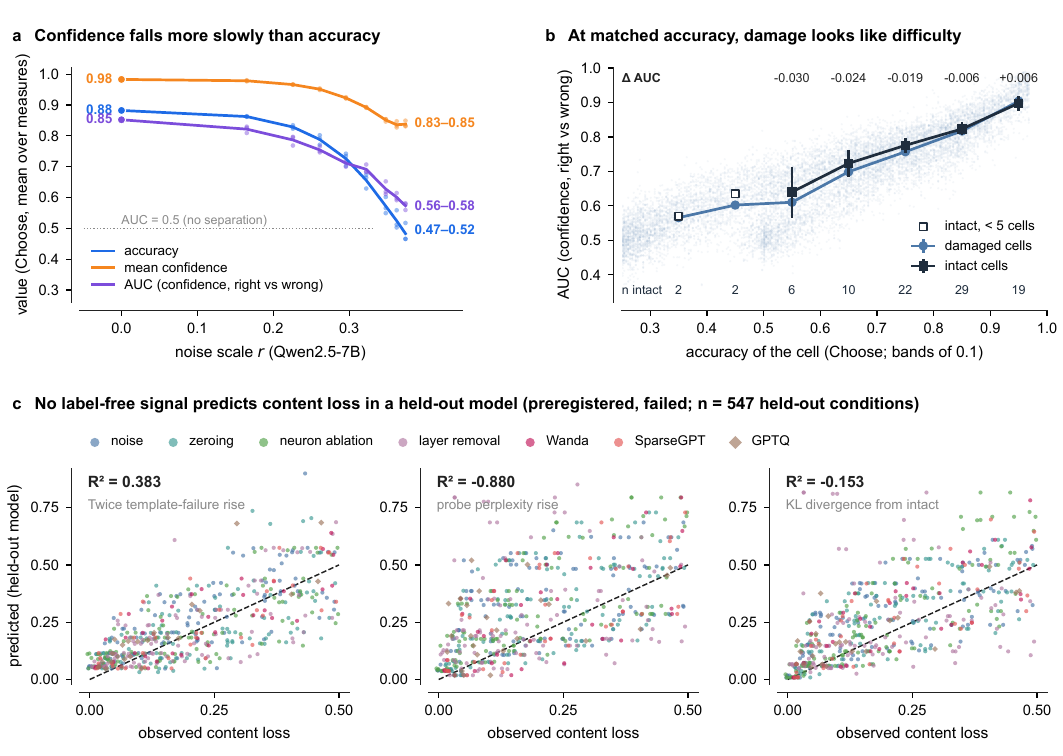}
\caption{\textbf{Confidence and label-free signals.} \textbf{a}, Qwen2.5-7B under noise: accuracy and mean confidence over 15 usable \cho{} measures, and the area under the curve (AUC) with which confidence separates right from wrong answers, against dose; small dots, seeds; lines, means (exploratory). \textbf{b}, AUC against accuracy for all damaged (circles) and intact (squares) cells of eight models, in accuracy bands of 0.1; numbers at the top, damaged minus intact; bars, 95\% intervals (exploratory). \textbf{c}, Held-out prediction of content loss from three label-free signals, the rise in \twc{} template failures, in perplexity on a probe text and in KL divergence from the intact model, each fitted on seven models and tested on the eighth ($n=547$ conditions; $R^2$ 0.383, $-0.880$ and $-0.153$; preregistered, failed).}
\label{fig:ed6}
\end{figure}
\clearpage

\section*{Supplementary Information (contents)}
Supplementary Table~1: the full audit of 20 studies. Supplementary Table~2: every preregistered test, threshold and outcome. Supplementary Table~3: instruction push per model. Supplementary Table~4: late-layer removal and random two-bit quantization. Supplementary Notes: protocol deviations; the four tests of intact-model confidence as accounts of the format effect; an earlier format battery; frequency counts and the reconciliation with studies that group facts by entity popularity; the separability of the fact and format axes; prediction of accuracy under compression; instruction push per model; late-layer removal and random two-bit quantization; further robustness checks.

\clearpage
\includepdf[pages=-]{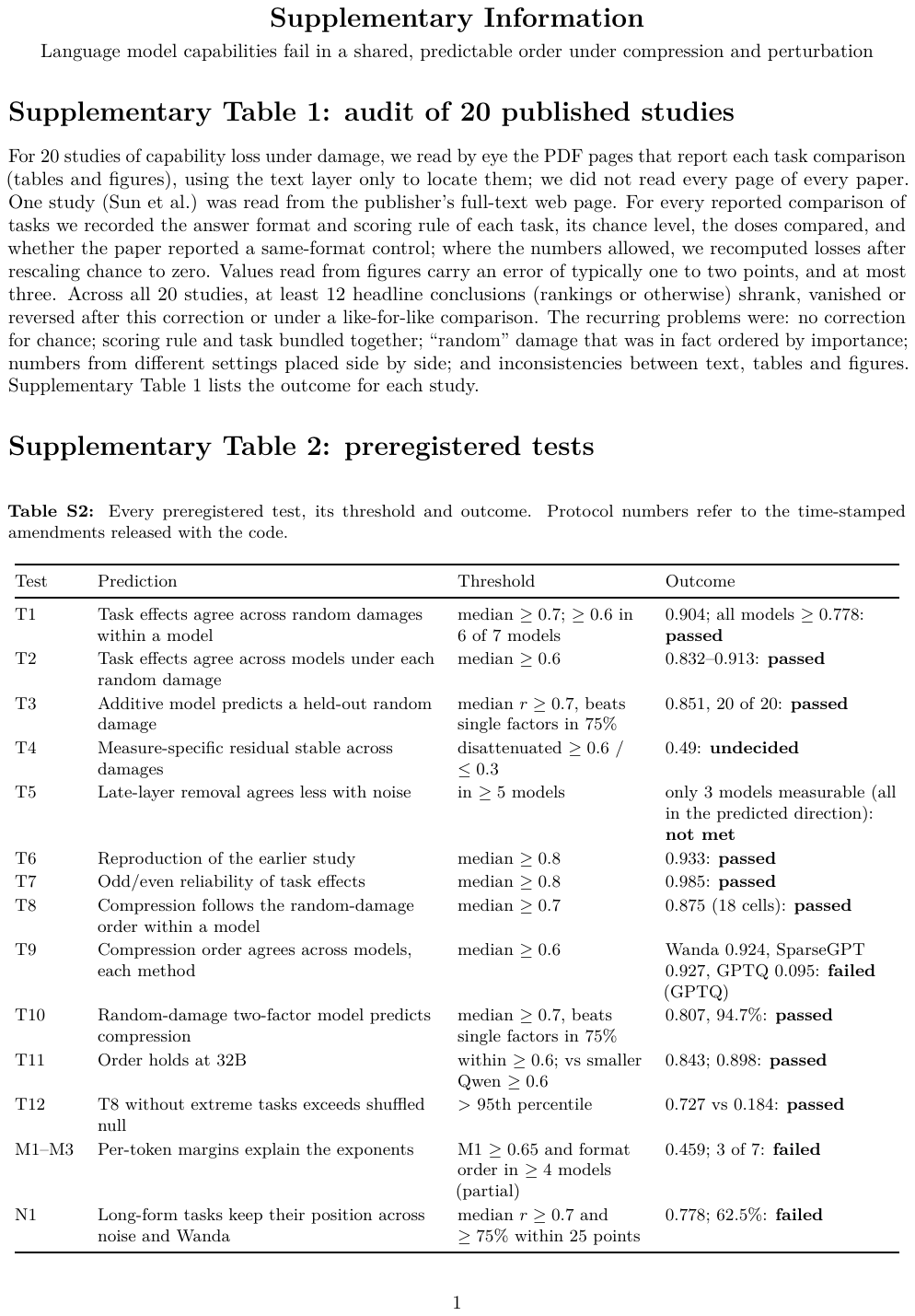}
\end{document}